%% file: main.tex
\documentclass[11pt, letterpaper, logo, onecolumn, copyright]{hunyuan_report}

\usepackage[authoryear, sort&compress, round]{natbib}
\usepackage{amssymb}
\usepackage{multirow}
\usepackage{enumitem}
\usepackage{booktabs}
\usepackage{graphicx}
\usepackage{wrapfig}
\usepackage{microtype}
\usepackage[breakable, skins]{tcolorbox}
\usepackage{colortbl}
\usepackage{amsmath}

\makeatletter
\def\input@path{{texmf/tex/latex/}}
\makeatother

\input{cross_document_refs.tex}

\newlength{\origcolwidth}
\newsavebox{\runningexamplebox}
\newenvironment{runningexample}{%
  \par\smallskip\noindent
  \begin{lrbox}{\runningexamplebox}%
  \begin{minipage}{\dimexpr\linewidth-2\fboxsep-2\fboxrule\relax}%
  \itshape
}{%
  \end{minipage}%
  \end{lrbox}%
  \fcolorbox{gray!45}{gray!8}{\usebox{\runningexamplebox}}%
  \par\smallskip
}

\title{Beyond Rephrasing: Book-Level Organization \\
Improves Synthetic Textbook Data for Mid-Training}
\runningtitle{Beyond Rephrasing: Book-Level Organization Improves Synthetic Textbook Data for Mid-Training}

\author{
  Jiawen Tao\textsuperscript{\rm 1,2}\textsuperscript{*},
  Miao Peng\textsuperscript{\rm 1,3}\textsuperscript{*},
  Yaoming Li\textsuperscript{\rm 2}\textsuperscript{*},
  Xiaokun Yuan\textsuperscript{\rm 1,2},
  Mengzhou Wu\textsuperscript{\rm 1,2}\\
  Wenhan Yu\textsuperscript{\rm 2},
  Guoan Wang\textsuperscript{\rm 1,2},
  Nuo Chen\textsuperscript{\rm 1}\textsuperscript{\textdagger},
  Tong Yang\textsuperscript{\rm 2}\textsuperscript{\textdagger},
  Maxm Pan\textsuperscript{\rm 1}\textsuperscript{\textdagger}\\
  \vspace{0.3cm}
  \normalsize
  \textsuperscript{1}Hunyuan Team, Tencent\\
  \textsuperscript{2}Peking University\\
  \textsuperscript{3}The Hong Kong University of Science and Technology (Guangzhou)\\
  {\fontfamily{lmtt}\selectfont taojiawen@stu.pku.edu.cn}\\
  {\fontfamily{lmtt}\selectfont yangtong@pku.edu.cn}\\
  {\fontfamily{lmtt}\selectfont \{maxmpan,kkelvinchen\}@tencent.com}
}

\begin{abstract}
Synthetic textbook data has improved language model pre-training, but prior work largely treats the benefit as a property of generated \emph{content} or local rewriting style. We study a different factor: whether related content is organized into coherent book-level documents. We contribute both a scalable synthesis pipeline and controlled evidence that this organization matters. The pipeline retrieves source material from a pre-training corpus, clusters it into topical units, plans hierarchical tables of contents, and assembles source-grounded sections into complete books (our Full setting), yielding 686K textbooks (32B tokens) across 15{,}000+ disciplines. Replacing natural books in a mid-training mix with this corpus improves downstream performance by $+1.09$ on average. Controlled comparisons then disentangle the relevant design factors. A content-matched Split condition holds generated text and tokens fixed but treats each section as an independent document; Full's $+1.02$ mean gain isolates \emph{document packaging}. A length-matched RandomConcat control that joins sections from different books remains below Full, ruling out document length alone. A retrieval-pool-matched Rephrase condition independently rewrites individual retrieved documents under the same audience$\times$style scheme, without clustering, TOC planning, or book assembly; Full's $+1.17$ gain demonstrates the value of \emph{structured synthesis}. On Llama3-8B, Full likewise outperforms both RandomConcat and Natural Books, supporting book-level organization as a useful axis for synthetic pre-training data design.
\end{abstract}

\begin{document}

\maketitle
\begingroup
\renewcommand{\thefootnote}{*}
\footnotetext{Equal contribution.}
\renewcommand{\thefootnote}{\textdagger}
\footnotetext{Corresponding authors.}
\endgroup
\setcounter{footnote}{0}

\section{Introduction}
\label{sec:intro}

\input{section1_intro.tex}

\section{Related Work}
\label{sec:related}

\input{section2_related.tex}

\input{section3_method.tex}

\section{Experimental Setup}
\label{sec:setup}

\input{section4_setup.tex}

\section{Results}
\label{sec:results}

\input{section5_results.tex}

\section{Ablation and Analysis}
\label{sec:ablation}

\input{section6_ablation_chunks.tex}

\input{section6_ablation_retrieval.tex}

\section{Conclusion}
\label{sec:conclusion}

\input{section7_conclusion.tex}

\bibliography{references}

\newpage
\appendix
\input{appendix_prompts.tex}

\end{document}

%% file: cross_document_refs.tex
\newcommand{\MainMethodSection}{\ref{sec:method}}
\newcommand{\MainPassRateTable}{\ref{tab:pass_rate}}
\newcommand{\MainBenchmarkTable}{\ref{tab:benchmark}}
\newcommand{\MainLlamaTable}{\ref{tab:llama_transfer}}
\newcommand{\MainComponentSection}{\ref{sec:ablation:retrieval}}

\newcommand{\SuppBenchmarkAppendix}{\ref{app:benchmark_details}}
\newcommand{\SuppPromptAppendix}{\ref{app:prompts}}

\newcommand{\SuppQueryPilotAppendix}{\ref{app:query_form_pilot}}
\newcommand{\SuppRandomConcatAppendix}{\ref{app:random_concat}}
\newcommand{\SuppLlamaAppendix}{\ref{app:llama_transfer}}
\newcommand{\SuppBootstrapAppendix}{\ref{app:benchmark_bootstrap}}
\newcommand{\SuppComponentAppendix}{\ref{app:component_ablation}}
\newcommand{\SuppDecontamAppendix}{\ref{app:decontam}}
\newcommand{\SuppComponentTable}{\ref{tab:retrieval_ablation}}

%% file: section1_intro.tex

Large language models derive their capabilities primarily from the scale, quality, and structure of pre-training data. Prior work on synthetic textbooks has shown that pedagogically written training data can improve model capabilities~\cite{gunasekar2023textbooks,li2023textbooks2}. This line of work has mainly emphasized content, namely the knowledge and explanations written into the text, or local rewriting style. By contrast, the question of how that content is \emph{organized}, assembled into coherent books rather than left as a flat collection of passages, remains underexplored.

We argue that organization is not incidental. On the generation side, a book-level plan provides the scaffold needed to turn scattered retrieved materials into long, structured, and coherent textbook data rather than a collection of locally plausible passages. On the training side, preserving that structure determines whether planned adjacent sections remain in a shared document with continuous positions and shared intra-document attention, or are split into independent examples with resets between sections. Our central question is therefore whether \textbf{book-level organization} adds value both when \textbf{constructing} synthetic data and when \textbf{preserving} that data as training documents, beyond the content itself or the local cleanup obtained by rephrasing individual documents.

To study this question at scale and produce structured, high-quality, long-form textbook data for mid-training, we introduce a retrieval-grounded pipeline over a pre-training corpus. Guided by a discipline taxonomy covering 15{,}000+ domains, it retrieves and clusters source documents, plans hierarchical tables of contents (TOCs) with quality filtering, and assembles source-grounded sections into complete books. The resulting corpus contains 686K textbooks and 32B tokens; it serves both as a scalable mid-training resource and as the empirical basis for controlled experiments that isolate the value of book-level organization.

Replacing the natural-book portion of a mid-training mix with this corpus
improves the 28-benchmark mean by $+1.09$. To understand where this gain comes
from, we perform controlled comparisons that separately isolate training-time
document organization, document length, and structured synthesis. Our
\textbf{Full} setting retains each TOC-planned textbook as a complete training
document. We compare it against three controls:
\textbf{Split} holds Full's generated sections and
token budget fixed but treats each section independently, resetting positions
and intra-document attention;
\textbf{RandomConcat} joins these Split sections
across books to match Full's length distribution, separating document length
from planned adjacency; and
\textbf{Rephrase} independently rewrites individual
documents from the same retrieval pool under the same audience$\times$style
scheme, without clustering, TOC planning, or book assembly. Across these
comparisons, organizing related content into coherent books consistently
outperforms the alternatives, showing that book-level organization---not
merely rewritten content or longer documents---is central to the effectiveness
of synthetic textbooks for mid-training.

Our contributions are threefold:
\begin{itemize}
\item \textbf{Scalable retrieval-grounded textbook synthesis.}
We introduce a pipeline that organizes retrieved material into
TOC-planned, source-grounded books, producing 686K textbooks (32B tokens)
across 15{,}000+ disciplines.
\item \textbf{Preserving book structure benefits model training.}
Content- and length-matched controls separate planned document continuity from
generated text and document length, showing that preserving coherent book
boundaries benefits training across two model architectures.
\item \textbf{Structured synthesis outperforms local rewriting.}
A matched rewriting control shows gains from global book construction beyond
independent document rewriting, while component ablations identify retrieval,
cluster-informed planning, and hierarchical generation as complementary
design factors.
\end{itemize}

%% file: section2_related.tex

\paragraph{Synthetic Textbooks and Rewriting.}
Synthetic corpora such as TinyStories~\cite{eldan2023tinystories} and the ``Textbooks Are All You Need'' line of work~\cite{gunasekar2023textbooks,li2023textbooks2,javaheripi2023phi2,abdin2024phi3} showed the value of high-quality pedagogical data. Cosmopedia~\cite{benallal2024cosmopedia} scaled this direction with synthetic textbooks and blog posts generated from curated web seeds, while WRAP~\cite{maini2024rephrasing} rephrases existing documents into cleaner styles. More structured variants move beyond flat topic prompting: Cosmopedia v2 uses a BISAC-style topic taxonomy and web retrieval to broaden educational coverage~\cite{huggingface2024smollm}; LiteLong uses BISAC-guided topics and BM25 retrieval to construct long-context training sequences~\cite{jia2026litelong}; and ACER generates TOC-based textbooks plus Bloom-taxonomy QA curricula for targeted domain knowledge infusion~\cite{neema2025acer}. These works improve coverage, grounding, and pedagogy but typically produce seed-level documents, long-context concatenations, or domain-specific curricula. We instead construct clustered, TOC-planned books from retrieved corpus material and test whether preserving this organization matters. Our \emph{Rephrase} baseline (\S\ref{sec:setup}) distinguishes this from per-document rewriting by matching Full's retrieval pool and audience$\times$style variation while omitting clustering, TOC planning, and book assembly.

\paragraph{Data Curation, Grounding, and Mid-Training.}
Data curation methods improve existing corpora through filtering, deduplication, selection, or mixture weighting~\cite{lee2022deduplicating,penedo2024fineweb,soldaini2024dolma,li2024datacomp,xie2023doremi,zhang2025autonomous,wettig2024qurating}. Retrieval-augmented and retrieval-grounded methods use retrieved documents to improve factuality, domain adaptation, or supervision~\cite{guu2020realm,lewis2020rag,borgeaud2022retro,source2synth2024,zhang2024raft}. Mid-training and continued pre-training inject domain knowledge without full retraining~\cite{gururangan2020dontstop,mo2025midtraining,taylor2022galactica,cheng2024adaptllm}. Our work is complementary: it uses retrieval not only to select or ground examples, but to transform corpus material into long-form, book-organized training documents for mid-training.

\paragraph{Document Boundaries, Packing, and Sequence Composition.}
A separate line of work studies how training examples are assembled into sequences, independent of their content. In-Context Pretraining~\cite{shi2024incontext} packs related documents into the same context, SPLiCe uses retrieval to pack mutually relevant documents for long-context training~\cite{staniszewski2025splice}, best-fit packing reduces cross-document truncation~\cite{ding2024fewer}, and \citet{zhao2024analysing} analyze how intra-document masking and sequence composition affect pre-training. Our Full vs.\ Split comparison asks the analogous question for synthetic textbooks: holding content and tokens fixed, does preserving a book's planned sections as one document help beyond treating each section as an independent training example?

\paragraph{Positioning.}
Together, these lines motivate organization as a first-class data-design axis.
We construct retrieved, TOC-planned books and test planned book boundaries
rather than post-hoc sequence assembly.

%% file: section3_method.tex
\section{Method}
\label{sec:method}

\subsection{Overview}
\label{sec:method:overview}

\begin{figure*}[t]
\centering
\includegraphics[width=\textwidth]{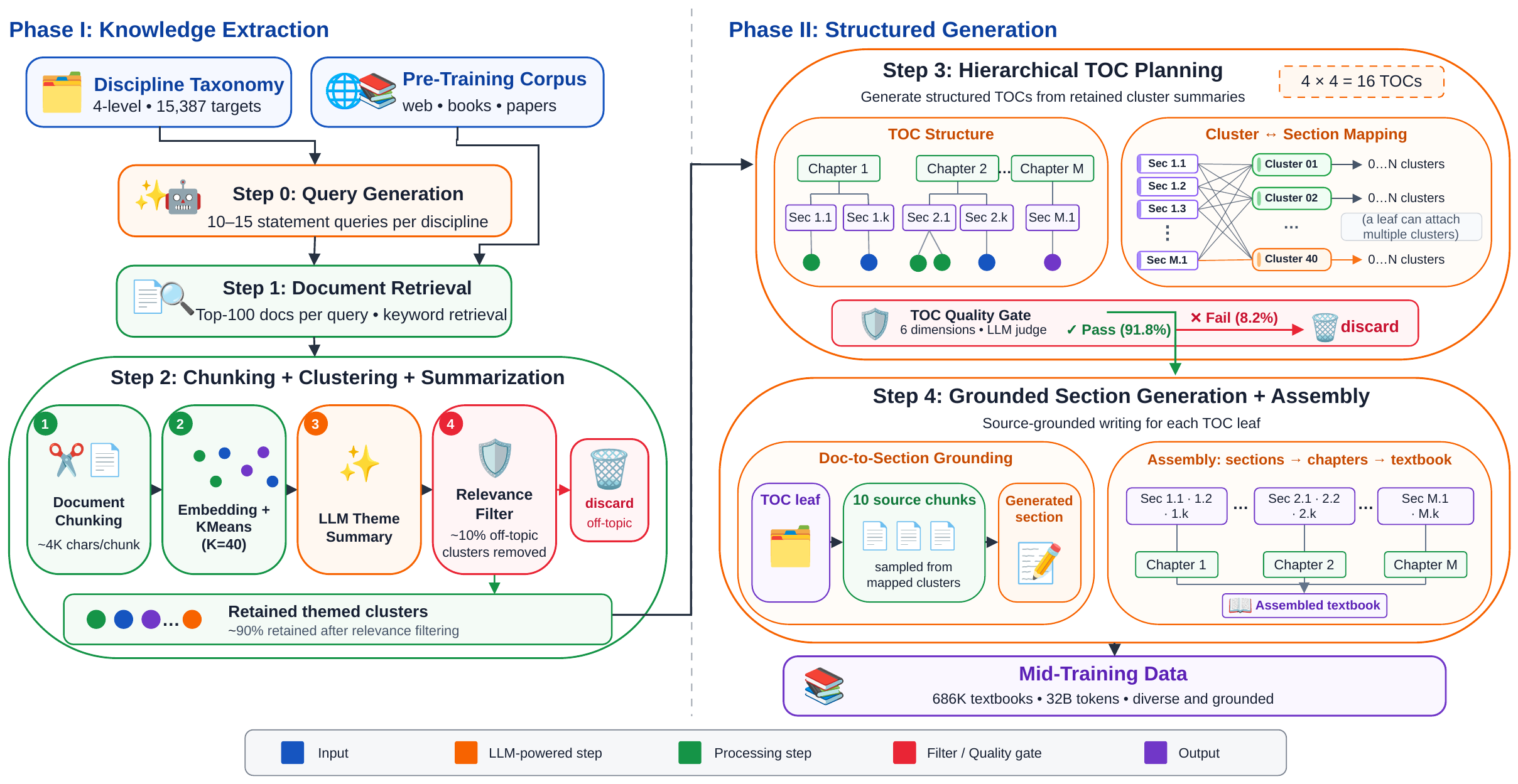}
\caption{Overview of the five-stage retrieval-grounded textbook synthesis pipeline. Phase~I (Steps~0--2) extracts and organizes knowledge from the pre-training corpus. Phase~II (Steps~3--4) plans structured TOCs and generates source-grounded content.}
\label{fig:pipeline}
\end{figure*}

We propose a five-stage pipeline that transforms raw pre-training corpora into structured, textbook-form training data. As illustrated in Figure~\ref{fig:pipeline}, the pipeline takes two inputs: (1) a discipline taxonomy covering 15,000+ knowledge domains, and (2) a pre-training corpus indexed for retrieval.

The pipeline is organized into two phases. \textbf{Phase~I} (Knowledge Extraction, Steps~0--2) generates diverse queries from the taxonomy, retrieves relevant documents, and organizes them into thematic clusters. \textbf{Phase~II} (Structured Generation, Steps~3--4) plans hierarchical table-of-contents structures, applies quality filtering, and generates source-grounded textbook content. The key design principles are: (i)~retrieval grounding at every generation stage to mitigate hallucination (\S\ref{sec:ablation:retrieval}); (ii)~hierarchical TOC planning for pedagogical structure; and (iii)~multi-variant generation across audiences and styles for distributional diversity.

\subsection{Discipline Taxonomy \& Query Generation}
\label{sec:method:step0}

\begin{wraptable}{r}{\origcolwidth}
\vspace{-4pt}
\centering
\footnotesize
\captionsetup{type=table,position=top,width=\linewidth,font=small,skip=12pt,justification=raggedright,singlelinecheck=false}
\setlength{\tabcolsep}{3pt}
\caption{Taxonomy coverage by domain.}
\label{tab:taxonomy}
\begin{tabular}{@{}l rrrr r@{}}
\toprule
\textbf{Domain} & \textbf{L1} & \textbf{L2} & \textbf{L3} & \textbf{L4} & \textbf{L3+L4} \\
\midrule
Natural Sciences         & 10 & 148 & 838  & 3,737  & 4,575 \\
Engineering \& Technology & 23 & 236 & 722  & 3,272  & 3,994 \\
Humanities \& Social Sci. & 19 & 251 & 857  & 4,090  & 4,947 \\
Medicine                 &  6 &  82 & 162  &   887  & 1,049 \\
Agriculture \& Life Sci. &  4 &  37 & 151  &   671  &   822 \\
\midrule
\textbf{Total}           & \textbf{62} & \textbf{754} & \textbf{2,730} & \textbf{12,657} & \textbf{15,387} \\
\bottomrule
\end{tabular}
\vspace{-2pt}
\end{wraptable}

\paragraph{Taxonomy.}
We construct a four-level discipline taxonomy for systematic knowledge coverage. The first three levels (62 L1 / 754 L2 / 2{,}730 L3 nodes) are adopted from the Chinese National Standard of Discipline Classification (GB/T~13745). An LLM then expands the L3 nodes into 12{,}657 finer-grained Level-4 subfields. To avoid inflated coverage from near-duplicate LLM expansions, we perform LLM-assisted deduplication under every L2 discipline before finalizing the L4 nodes. We use both the 2{,}730 L3 nodes and the 12{,}657 L4 nodes as generation targets, for a total of 15{,}387 disciplines (i.e., $\text{L3}+\text{L4}$; Table~\ref{tab:taxonomy}) spanning five broad domains.

\paragraph{Query Generation.}
For each leaf discipline, an LLM generates 10--15 diverse queries in \emph{statement form}, covering different knowledge angles (core concepts, methods, applications, limitations). We choose statement-form queries because they better match the declarative register of pre-training text and produced higher cluster-retention rates in pilot relevance checks (Appendix~\SuppQueryPilotAppendix{}).

\begin{runningexample}
\textbf{Running Example.} \emph{Feedforward Neural Networks} (CS $\to$ AI $\to$ Neural Networks) $\to$ 14 statement queries, e.g., ``The backpropagation algorithm efficiently computes gradients of a loss function with respect to all network parameters.''
\end{runningexample}

\subsection{Retrieval and Clustering}
\label{sec:method:step1}

Each query retrieves the top-100 passages from the pre-training corpus via keyword matching. Passages across all queries for a discipline are aggregated and deduplicated. The methodology is retrieval- and corpus-agnostic: dense retrievers or other indexed corpora can be substituted.

Retrieved documents are split into chunks, embedded with a text embedding model, and clustered via KMeans. Clustering turns a flat retrieval ranking into a \emph{structured knowledge map}: each cluster is a coherent subtopic that can map to chapters or sections, rather than an undifferentiated list ranked only by retrieval score.

For each cluster, representative documents (by centroid proximity) are sampled and an LLM generates a one-sentence theme summary. Simultaneously, the LLM judges whether the cluster falls within the discipline's scope; off-topic clusters are excluded from subsequent TOC planning.

\begin{runningexample}
\textbf{Running Example.} 1,346 unique documents retrieved $\to$ 40 clusters. Themes include ``Universal Approximation Theory'', ``SGD and Learning Rate Scheduling'', ``Physics-Informed Neural Network Training''.
\end{runningexample}

\subsection{TOC Planning with Quality Gate}
\label{sec:method:step3}

\paragraph{TOC Generation.}
Given the filtered cluster map, the planner converts a set of retrieved subtopics into an explicit book blueprint rather than a flat topic list. Its input contains the discipline path, the target audience and writing style, the global length budget, and for each retained cluster: a theme summary, representative evidence snippets, and its in-scope judgment. The LLM must output a machine-readable hierarchical TOC (chapters $\rightarrow$ sections $\rightarrow$ leaf sections). Each leaf specifies (i) a title and short learning objective, (ii) the source cluster IDs it draws from, and (iii) a token-budget estimate used later by the section generator. This makes the TOC a routing structure: downstream generation can retrieve the exact source chunks associated with each leaf, and the assembly step can preserve the planned chapter/section order.

We generate TOCs for 4 audiences $\times$ 4 styles = 16 variants per discipline. The variants are not simple paraphrases: audiences select different levels of abstraction from the retrieved reference pool (e.g., intuitive motivation for high-school readers versus formal definitions and frontier issues for researchers), while styles induce different organizational skills such as exposition, procedural tutoring, encyclopedic coverage, and practitioner-oriented case analysis. This produces distributional diversity while keeping all variants anchored to the same retrieved knowledge map.

\paragraph{Quality Gate.}
Before generating any section text, we apply an LLM-based quality gate to each candidate TOC. The judge evaluates whether the plan is coherent, appropriate for the requested audience/style, sufficiently grounded in the retained source clusters, and ready for section-level generation. TOCs that fail this gate are discarded, since structural defects would otherwise propagate to every descendant section and are cheaper to catch at the planning stage than after full-book generation. Corpus-level pass rates and their audience/style variation are reported in \S\ref{sec:setup} (Table~\ref{tab:pass_rate}).

\begin{runningexample}
\textbf{Running Example.} 15 of 16 TOC variants pass quality gate. College/textbook variant: 4 chapters, 16 sections, referencing 26/35 clusters.
\end{runningexample}

\subsection{Source-Grounded Generation \& Assembly}
\label{sec:method:step4}

For each TOC leaf node, several source chunks are sampled from its mapped clusters and provided as context to the LLM. The model writes section content \emph{conditioned on retrieved evidence}, ensuring each section is grounded in corpus material rather than relying solely on parametric knowledge. This section-by-section approach maintains retrieval anchoring throughout long-form generation.

Sections are assembled into complete books following the TOC hierarchy. For each passed TOC variant, we generate multiple independent books by resampling source chunks, yielding many textbook variants per discipline under the same global plan and audience/style specification.

\begin{runningexample}
\textbf{Running Example.} 45 books generated from 15 passed TOCs ($\times$3 copies), comprising 630 sections and 10.5M characters. Sample edition: 4 chapters, 16 sections, 257K characters.
\end{runningexample}

%% file: section4_setup.tex

\paragraph{Model and Training.}
We train five matched variants of the same in-house mixture-of-experts (MoE)
language model with 3B active parameters (30B total). Each uses the same
200B-token mid-training mix (one epoch) spanning web, math, papers, books, and
other sources. Optimizer, schedule, batch size, tokenizer, and data order are
held fixed; the only manipulated component is the book portion of the mix
(16B tokens, 8\%). This fixed allocation matches the original book slice,
providing a practical replacement setting while keeping the overall mixture
and token budget unchanged; we do not optimize the synthetic-data proportion.
As a transfer check, we also mid-train Llama3-8B under three book conditions
(Full, RandomConcat, Natural Books). Each uses the same 100B-token mixture with
an 8B-token book slice (8\%); data order, training steps, optimizer, and
learning-rate schedule are held fixed, and only the book condition changes.
Detailed hyperparameters are in Appendix~\SuppLlamaAppendix{}. The five primary
MoE runs and three Llama runs require approximately 89.6K and 39.9K H800
GPU-hours, respectively.

\paragraph{Pipeline components.}
Retrieval uses keyword matching over an in-house pre-training corpus (top-100 passages per query). Retrieved passages are chunked ($\sim$4K characters), embedded with OpenAI's text-embedding-3-small, and clustered with KMeans ($K{=}40$). Query generation, TOC planning, and the quality gate use DeepSeek-V3.2; Qwen3.5-35B-A3B generates sections from $k{=}10$ source chunks (\S\ref{sec:ablation:chunks}). Intrinsic-quality ablations use Gemini-3.1-Pro as an independent judge; it does not generate or filter training data. Key parameters were fixed in pilots over 50 disciplines; full prompts are in the appendix.

\paragraph{Corpus Statistics.}
The pipeline generates 249K TOCs, of which 91.8\% pass the quality gate
(Table~\ref{tab:pass_rate}); high-school variants have the lowest pass rate,
reflecting the difficulty of simplifying university-level source material.
From passed plans we produce 686K books totaling 32B tokens, with up to 48
variants per discipline (16 audience$\times$style TOCs $\times$ 3 resampled
books). Final text generation costs 20K H20 GPU-hours for Full versus 19.7K
for Rephrase; Full adds about 1K for cluster summarization, with other upstream
costs being small relative to final generation.

\begin{wraptable}{r}{\origcolwidth}
\vspace{-4pt}
\centering
\footnotesize
\captionsetup{type=table,position=top,width=\linewidth,font=small,skip=12pt,justification=raggedright,singlelinecheck=false}
\setlength{\tabcolsep}{3.5pt}
\caption{TOC quality gate pass rate (\%) by audience $\times$ style.}
\label{tab:pass_rate}
\begin{tabular}{@{}l cccc c@{}}
\toprule
\textbf{Audience} & \textbf{Blog} & \textbf{Textbook} & \textbf{Tutorial} & \textbf{Wiki} & \textbf{Avg} \\
\midrule
High School  & 90.0 & 72.9 & 80.2 & 60.2 & 75.8 \\
College      & 98.6 & 97.8 & 96.2 & 96.8 & 97.3 \\
Researcher   & 98.1 & 97.8 & 93.9 & 97.4 & 96.8 \\
Practitioner & 98.2 & 96.5 & 97.5 & 95.9 & 97.0 \\
\midrule
\textbf{Avg} & 96.2 & 91.3 & 92.0 & 87.6 & \textbf{91.8} \\
\bottomrule
\end{tabular}
\vspace{-2pt}
\end{wraptable}

\begin{figure*}[t]
\centering
\includegraphics[width=\textwidth]{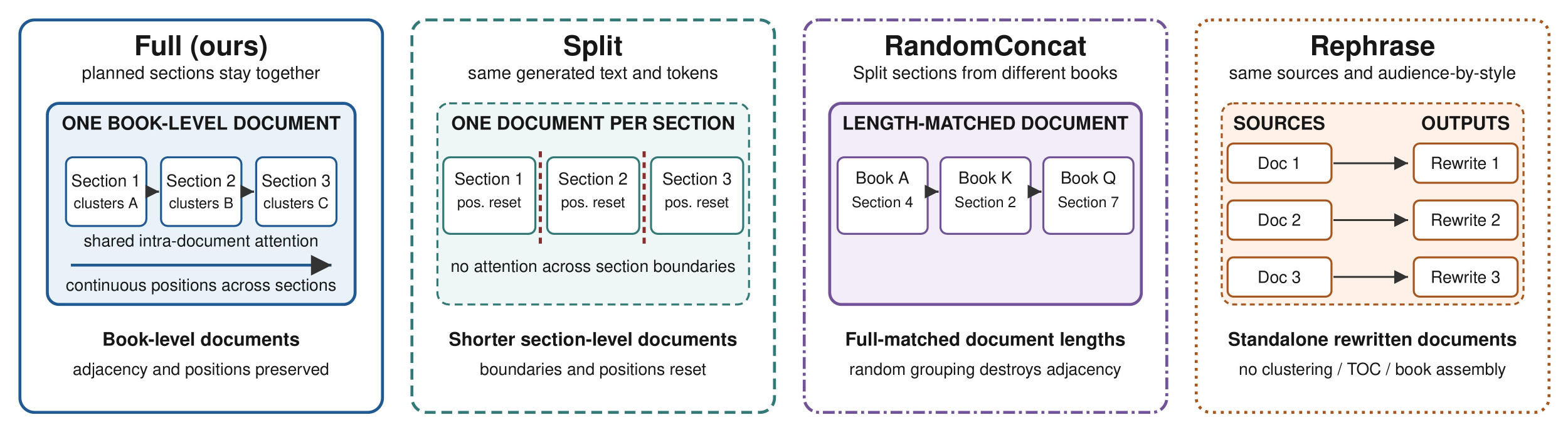}
\caption{Training conditions. Split holds generated text and tokens fixed;
RandomConcat matches Full's document lengths using sections from different
books; Rephrase independently rewrites documents from the same retrieval pool; Natural Books is the
original-mix reference.}
\label{fig:controls}
\end{figure*}

\textbf{Full} (ours) replaces the book slice with textbooks from our 32B-token corpus and keeps each book as one document. \textbf{Split} uses identical text and tokens but makes each section independent. \textbf{RandomConcat} groups sections from different books to match Full's document-length distribution. \textbf{Rephrase} independently rewrites documents from the same retrieval pool under the same audience$\times$style scheme, without clustering, TOC planning, or book assembly, under the same 16B-token budget. \textbf{Natural Books} retains the original book slice as a practical, unmatched baseline. Together, these controls test packaging, planned adjacency, structured synthesis, and practical replacement (Figure~\ref{fig:controls}).

\paragraph{Sequence packing.}
All conditions use sequence packing with \emph{intra-document attention masking}: tokens attend only to tokens from the same document, and both attention and position indices are reset at document boundaries.

\begin{wrapfigure}[16]{r}{\origcolwidth}
\vspace{-4pt}
\centering
\captionsetup{type=figure,position=bottom,width=\linewidth,font=footnotesize,skip=8pt}
\includegraphics[width=\linewidth]{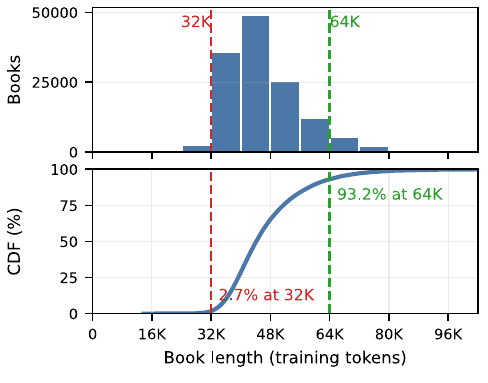}
\caption{Book-length histogram and CDF from a 20\% per-shard sample.}
\label{fig:lengthdist}
\vspace{-2pt}
\end{wrapfigure}

\paragraph{Book lengths.}
TOC planning targets a book-scale budget of about 50K training tokens
($\approx$37K English words under the prompt conversion of 0.75 words per
token). Figure~\ref{fig:lengthdist} shows that realized books concentrate near
this scale (median 44.2K; 90\% fall between 32K and 64K).

\paragraph{Decontamination.}
We remove generated books flagged by mixed Chinese/English 13-gram overlap
with benchmark tests (Appendix~\SuppDecontamAppendix{}). Only ${\sim}$0.07\%
of tokens are flagged in either retrieved sources or generated books, making
verbatim leakage unlikely to explain the gains.

\paragraph{Evaluation.}
We evaluate on 28 benchmarks spanning STEM~\citep{hendrycks2021math,shi2023mgsm,rein2024gpqa,ma2025korbench}, knowledge and multilingual understanding~\citep{hendrycks2021mmlu,wang2024mmlupro,huang2023ceval,pteam2025supergpqa,kwiatkowski2019naturalquestions,zhong2024agieval,wei2024simpleqa,goyal2022flores,bandarkar2024belebele}, reasoning and mathematical word-problem solving~\citep{zellers2019hellaswag,sap2019socialiqa,dua2019drop,suzgun2023bbh,cobbe2021gsm8k}, and code~\citep{chen2021codex,austin2021program,liu2023evalplus,jain2025livecodebench,zhuo2025bigcodebench} (Table~\ref{tab:benchmark}). All conditions use identical evaluation settings in a unified harness: multiple-choice tasks use few-shot likelihood scoring, while generative tasks use greedy decoding (temperature 0), except where the benchmark's standard protocol requires sampling. Protocol details are provided in Appendix~\SuppBenchmarkAppendix{}.

%% file: section5_results.tex

We report two matched mid-training studies: five book conditions on a
3B-active MoE (Table~\ref{tab:benchmark}) and three on Llama3-8B
(Table~\ref{tab:llama_transfer}). The latter tests transfer across architectures.

\begin{table}[!t]
\centering
\caption{Downstream benchmark results at a fixed mid-training checkpoint. $\Delta_{\mathrm{S}}$, $\Delta_{\mathrm{R}}$, and $\Delta_{\mathrm{N}}$ denote Full minus Split, Rephrase, and Natural Books, respectively. Bold marks the best among Full, Split, and Rephrase per benchmark (Natural Books is compared separately via $\Delta_{\mathrm{N}}$ and is never bolded); italic rows report unweighted category means.}
\label{tab:benchmark}
\resizebox{\linewidth}{!}{%
\small
\setlength{\tabcolsep}{5pt}%
\begin{tabular}{ll cccc ccc}
\toprule
\textbf{Category} & \textbf{Benchmark} & \multicolumn{4}{c}{\textbf{Scores}} & \multicolumn{3}{c}{\textbf{Deltas}} \\
\cmidrule(lr){3-6}\cmidrule(lr){7-9}
 & & \textbf{Full (ours)} & \textbf{Split} & \textbf{Rephrase} & \textbf{Natural Books} & \textbf{$\Delta_{\mathrm{S}}$} & \textbf{$\Delta_{\mathrm{R}}$} & \textbf{$\Delta_{\mathrm{N}}$} \\
\midrule
\multirow{7}{*}{STEM}
  & CMATH             & \textbf{85.17} & 85.00 & 83.50 & 82.83 & +0.17 & +1.67 & +2.34 \\
  & MATH              & 50.10 & 49.72 & \textbf{50.20} & 50.65 & +0.38 & $-$0.10 & $-$0.55 \\
  & U-MATH            & \textbf{37.43} & 35.79 & 36.54 & 37.32 & +1.64 & +0.89 & +0.11 \\
  & MGSM              & \textbf{69.24} & 68.91 & 68.65 & 69.45 & +0.33 & +0.59 & $-$0.21 \\
  & GPQA-Diamond      & \textbf{33.20} & 30.05 & 32.79 & 26.29 & +3.15 & +0.41 & +6.91 \\
  & UGPhysics         & 19.62 & \textbf{19.76} & 18.80 & 19.49 & $-$0.14 & +0.82 & +0.13 \\
  & KORBench          & 40.24 & \textbf{41.84} & 35.36 & 35.92 & $-$1.60 & +4.88 & +4.32 \\
\cmidrule(l){2-9}
  & \emph{STEM mean}  & \textbf{47.86} & 47.30 & 46.55 & 45.99 & +0.56 & +1.31 & +1.86 \\
\midrule
\multirow{11}{*}{Knowledge}
  & MMLU              & 76.18 & \textbf{76.24} & 76.09 & 76.76 & $-$0.06 & +0.09 & $-$0.58 \\
  & MMLU-Redux        & \textbf{77.28} & 76.91 & 76.96 & 77.81 & +0.37 & +0.32 & $-$0.53 \\
  & MMLU-Pro          & \textbf{49.65} & 49.07 & 49.13 & 49.07 & +0.58 & +0.52 & +0.58 \\
  & C-Eval            & \textbf{81.20} & 81.13 & 80.24 & 80.68 & +0.07 & +0.96 & +0.52 \\
  & SuperGPQA         & 32.51 & 32.37 & \textbf{33.22} & 33.11 & +0.14 & $-$0.71 & $-$0.60 \\
  & NaturalQuestions  & \textbf{35.40} & 34.88 & 32.47 & 35.18 & +0.52 & +2.93 & +0.22 \\
  & AGIEval-EN        & \textbf{55.13} & 54.26 & 54.51 & 53.89 & +0.87 & +0.62 & +1.24 \\
  & INCLUDE-Base-44   & \textbf{71.34} & 70.66 & 70.13 & 71.05 & +0.68 & +1.21 & +0.29 \\
  & SimpleQA          & \textbf{9.00} & 8.54 & 8.45 & 8.59 & +0.46 & +0.55 & +0.41 \\
  & FLORES            & \textbf{26.88} & 24.24 & 26.33 & 24.05 & +2.64 & +0.55 & +2.83 \\
  & Belebele          & \textbf{90.82} & 90.41 & 90.21 & 90.48 & +0.41 & +0.61 & +0.34 \\
\cmidrule(l){2-9}
  & \emph{Knowledge mean} & \textbf{55.04} & 54.43 & 54.34 & 54.61 & +0.61 & +0.70 & +0.43 \\
\midrule
\multirow{5}{*}{Reasoning}
  & HellaSwag         & 81.23 & \textbf{81.30} & 80.67 & 81.37 & $-$0.07 & +0.56 & $-$0.14 \\
  & SIQA              & \textbf{75.90} & 74.67 & 74.62 & 73.85 & +1.23 & +1.28 & +2.05 \\
  & DROP              & 72.20 & 72.50 & \textbf{73.37} & 72.40 & $-$0.30 & $-$1.17 & $-$0.20 \\
  & BBH               & 76.53 & \textbf{77.03} & 76.18 & 76.33 & $-$0.50 & +0.35 & +0.20 \\
  & GSM8K             & \textbf{88.02} & 81.64 & 83.70 & 82.49 & +6.38 & +4.32 & +5.53 \\
\cmidrule(l){2-9}
  & \emph{Reasoning mean} & \textbf{78.78} & 77.43 & 77.71 & 77.29 & +1.35 & +1.07 & +1.49 \\
\midrule
\multirow{5}{*}{Code}
  & HumanEval+        & \textbf{69.75} & 64.60 & 67.08 & 65.84 & +5.15 & +2.67 & +3.91 \\
  & MBPP+             & \textbf{69.07} & 68.00 & 67.11 & 66.31 & +1.07 & +1.96 & +2.76 \\
  & LiveCodeBench     & \textbf{25.65} & 25.50 & 24.27 & 25.88 & +0.15 & +1.38 & $-$0.23 \\
  & BigCodeBench      & \textbf{46.84} & 45.44 & 45.61 & 46.58 & +1.40 & +1.23 & +0.26 \\
  & BigCodeBench-Hard & \textbf{19.59} & 16.22 & 16.22 & 20.95 & +3.37 & +3.37 & $-$1.36 \\
\cmidrule(l){2-9}
  & \emph{Code mean}  & \textbf{46.18} & 43.95 & 44.06 & 45.11 & +2.23 & +2.12 & +1.07 \\
\midrule
\textbf{Overall (28)} & & \textbf{55.90} & 54.88 & 54.73 & 54.81 & +1.02 & +1.17 & +1.09 \\
\bottomrule
\end{tabular}%
}
\end{table}

\paragraph{Book boundaries help beyond identical content.}
Split resets positions and intra-document attention at each section;
RandomConcat restores Full-matched lengths by random grouping but destroys
planned coherence.
Full exceeds content-identical Split on 22/28 benchmarks ($+1.02$), leading
every category mean (code $+2.23$, reasoning $+1.35$, knowledge $+0.61$, STEM
$+0.56$). Length-matched RandomConcat ties Split (54.89 vs.\ 54.88) but
trails Full by $1.01$ (21/28 wins; Appendix~\SuppRandomConcatAppendix{}),
supporting planned adjacency beyond length matching and fewer resets.

\paragraph{The structured pipeline outperforms simple rephrasing.}
At comparable generation and identical training compute, Full beats Rephrase
on 25/28 benchmarks ($+1.17$), while Rephrase ties Natural Books ($-0.08$).
Unlike Split and RandomConcat, this comparison changes local text while
matching the source pool and style distribution, thereby isolating
generation-side organization from training-time packaging.
Rephrase's near-equality to Natural Books indicates that local rewriting alone
is insufficient; Full's additional gain instead supports clustering, TOC
planning, and book assembly.

\paragraph{Gains span all four categories.}
Full improves all four category means over Natural Books---STEM ($+1.86$),
reasoning ($+1.49$), code ($+1.07$), and knowledge ($+0.43$)---and 19/28
benchmarks ($+1.09$ overall). The larger STEM, reasoning, and code gains are
consistent with textbook-form data emphasizing structured procedural and
disciplinary exposition over factoid recall.

Benchmark-level bootstrap intervals remain positive for all three aggregate
comparisons. One-sided sign tests give $p{=}0.044$, $0.0019$, and $0.000014$
for Full against Natural Books, Split, and Rephrase, respectively. Together,
these checks show that the aggregate gains are not driven by a few outlier
benchmarks, while not estimating training-run variance
(Appendix~\SuppBootstrapAppendix{}).

\paragraph{Multi-model validation.}
\begin{wraptable}[12]{r}{0.48\textwidth}
\vspace{-11pt}
\centering
\footnotesize
\captionsetup{type=table,position=top,width=\linewidth,font=footnotesize,skip=6pt,justification=raggedright,singlelinecheck=false}
\setlength{\tabcolsep}{2.5pt}
\caption{Llama3-8B means. $\Delta_{\mathrm{F-R}}=$ Full$-$RandomConcat;
$\Delta_{\mathrm{F-N}}=$ Full$-$Natural Books.}
\label{tab:llama_transfer}
\begin{tabular}{@{}l ccccc@{}}
\toprule
\textbf{Category} & \textbf{Full} &
\textbf{\shortstack{Random-\\Concat}} &
\textbf{\shortstack{Natural\\Books}} &
\textbf{$\Delta_{\mathrm{F-R}}$} &
\textbf{$\Delta_{\mathrm{F-N}}$} \\
\midrule
STEM      & \textbf{27.35} & 25.86 & 23.70 & +1.49 & +3.65 \\
Knowledge & \textbf{42.42} & 42.17 & 41.90 & +0.24 & +0.52 \\
Reasoning & \textbf{68.83} & 67.67 & 67.63 & +1.17 & +1.20 \\
Code      & \textbf{28.94} & 27.92 & 27.89 & +1.02 & +1.05 \\
\midrule
Overall (28) & \textbf{40.96} & 40.10 & 39.45 & +0.86 & +1.51 \\
\bottomrule
\end{tabular}
\vspace{-4pt}
\end{wraptable}
On Llama3-8B, Full scores 40.96, versus 40.10 for RandomConcat and 39.45 for
Natural Books ($+0.86$; 20/28 wins; 95\% CI $[+0.44,+1.30]$; $p{=}0.018$;
Table~\ref{tab:llama_transfer}). The consistent ordering supports planned
section order beyond length matching in a second architecture. The gain over
RandomConcat is positive in every category, from $+0.24$ in knowledge to
$+1.49$ in STEM; Appendix~\SuppLlamaAppendix{} provides the complete training
recipe and benchmark-level robustness results.
\WFclear
\input{table_chunk_ablation.tex}

%% file: table_chunk_ablation.tex
\begin{table}[t]
\centering
\small
\caption{Effect of source chunk count ($k$) on section quality (1--5 scale).
Input length in characters. Shaded row: selected operating point ($k{=}10$).}
\label{tab:chunk_ablation}
\begin{tabular}{r r cccccccc}
\toprule
$k$ & \textbf{Input} & \textbf{Overall} & \textbf{Factual} & \textbf{Theme} & \textbf{Structure} & \textbf{Edu.} & \textbf{Aud.Fit} & \textbf{Breadth} & \textbf{Depth} \\
\midrule
1  & 12.8K & 4.548 & 4.753 & 4.793 & 4.688 & 4.740 & 4.860 & 4.855 & 4.880 \\
3  & 20.1K & 4.550 & 4.683 & 4.707 & 4.647 & 4.665 & 4.835 & 4.867 & 4.870 \\
5  & 27.7K & 4.643 & 4.760 & 4.782 & 4.700 & 4.772 & 4.903 & 4.955 & 4.952 \\
\rowcolor{gray!15}
10 & 46.5K & 4.768 & 4.790 & 4.893 & 4.850 & 4.872 & 4.935 & 4.975 & 4.963 \\
20 & 84.0K & 4.777 & 4.808 & 4.920 & 4.853 & 4.885 & 4.940 & 4.980 & 4.978 \\
30 & 121.5K & 4.822 & 4.838 & 4.942 & 4.893 & 4.918 & 4.955 & 4.987 & 4.982 \\
\bottomrule
\end{tabular}
\end{table}

%% file: section6_ablation_chunks.tex
\subsection{Effect of Source Chunk Count}
\label{sec:ablation:chunks}

In Step~4, each section is generated conditioned on $k$ source chunks sampled from its mapped clusters. We vary $k \in \{1, 3, 5, 10, 20, 30\}$ and evaluate generation quality on 400 individual sections using Gemini-3.1-Pro as the LLM judge, scoring seven dimensions plus overall on a fixed 1--5 rubric. Since chunk count directly affects per-section generation, we evaluate at the section level rather than full-book level. Definitions and score anchors are provided in Appendix~\SuppPromptAppendix{}.

From $k{=}1$ to $k{=}10$, overall, structure, and educational scores improve by
$+0.22$, $+0.16$, and $+0.13$. From 10 to 30 chunks, overall improves only
$+0.05$ while breadth and depth change by $<0.02$; $k{=}20$ nearly doubles
input for a $+0.009$ overall gain. We therefore select $k{=}10$, which captures
most of the benefit at ${\sim}$47K input characters per section.

%% file: section6_ablation_retrieval.tex
\subsection{Retrieval Grounding and Pipeline Components}
\label{sec:ablation:retrieval}

We compare the full system with three progressively reduced baselines:
\textbf{A} keeps the identical cluster-informed TOC but removes source chunks;
\textbf{B} also removes cluster-informed planning by generating a TOC from the
discipline name alone; and \textbf{C} directly generates an entire book without
TOC planning, clustering, or source chunks.

On 100 matched book instances with aligned audience and style,
Gemini-3.1-Pro scores each complete book on factuality, specificity, breadth,
depth, structure, pedagogy, originality, example quality, and audience/style
fit, plus an overall score, on a fixed 1--10 rubric. The same judge and rubric
are used across conditions; Baseline~A retains Full's TOC. These intrinsic
scores diagnose component-level effects rather than replace the downstream
evaluations in Table~\ref{tab:benchmark}.
Definitions and score anchors are in Appendix~\SuppPromptAppendix{}, and full
results are in Appendix~\SuppComponentAppendix{}
(Table~\SuppComponentTable{}).

\paragraph{Structured generation has the largest observed effect.}
Baseline~C scores 6.12 overall versus 9.38 for Full ($-3.26$). Despite a
${\sim}$30--40K-word length target, it still yields only ${\sim}$56K characters
(roughly 4$\times$ shorter than Full), with lower depth (5.86) and specificity
(6.41). This supports hierarchical TOC planning plus section-by-section
generation for long-form data with sustained coverage and depth.

\paragraph{Source chunks primarily improve specificity and examples.}
Full versus Baseline~A isolates retrieval grounding: source chunks improve
specificity ($+0.23$), examples ($+0.35$), and factual accuracy ($+0.32$),
with smaller differences elsewhere. Grounding therefore mainly contributes
domain-specific details rather than structure or pedagogy.

\paragraph{Cluster-informed TOC improves structure and audience fit.}
A uses Full's cluster-informed TOC; B generates one from the discipline name.
Neither uses source chunks, isolating cluster-informed planning. A improves
audience/style fit (9.21 vs.\ 8.62, $+0.59$), structure (9.35 vs.\ 8.91,
$+0.44$), pedagogy (9.23 vs.\ 9.00, $+0.23$), and overall quality
(9.21 vs.\ 9.07, $+0.14$), showing that source coverage supports more targeted
TOCs before any section text is generated.

\paragraph{TOC filtering acts as a safety net.}
As described in \S\ref{sec:method:step3}, the judge scores 6 structural
dimensions and rejects 8.2\% of TOCs. We generate books from 200 failed TOCs
(Baseline~D) and evaluate them with the same judge. Relative to Full, they
score lower on overall ($-0.25$), structure ($-0.20$), and pedagogy
($-0.19$). This smaller degradation than removing source chunks or cluster
summaries makes the gate a safety net rather than the primary quality driver.

\paragraph{The components play complementary roles.}
Across these ablations, retrieval grounding chiefly adds domain-specific
details, cluster-informed planning improves structure and audience fit, and
hierarchical section-wise generation supplies book-scale depth and coverage.
Global organization and local grounding are therefore complementary rather
than interchangeable across stages of synthesis.

%% file: section7_conclusion.tex

In this work, we propose a retrieval-grounded synthesis pipeline that scales to
686K books (32B tokens) across 15{,}000+ disciplines, and use the resulting
corpus to systematically study book-level organization in both textbook
construction and mid-training. Under matched training conditions within each
study, Full ranks first on both the 3B-active MoE and Llama3-8B. The controlled
comparisons explain this result through two complementary effects. On the
generation side, clustering and hierarchical TOC planning determine which
knowledge is combined and how it progresses, while book assembly realizes this
plan as a coherent document. The Rephrase comparison shows that this global
synthesis adds value beyond local rewriting. On the training side, preserving
the resulting book structure is separately important: Split removes document
continuity, while RandomConcat matches Full's document-length distribution but
destroys semantic order. Their consistent gaps from Full suggest that effective
textbook data requires both constructing coherent books and exposing their
planned adjacency during training. Component ablations further identify structured generation as
the primary driver, with retrieval, cluster-informed TOCs, and quality filtering
providing complementary gains. We provide synthesis and control-construction
code as supplementary material and will release a research-licensed subset of
the synthetic corpus upon publication.

\section{Limitations}
Although the effect transfers across two model settings, each condition is
trained once, so training-run variance remains unmeasured. The pipeline also
requires a searchable corpus index, which adds preprocessing and
retrieval-infrastructure overhead. Finally, component ablations rely on a fixed
LLM judge and should be interpreted as diagnostic rather than downstream
evidence.

%% file: appendix_prompts.tex
\section{Benchmark Details}
\label{app:benchmark_details}

Table~\MainBenchmarkTable{} reports results on 28 benchmarks grouped into four categories: \emph{STEM} (CMATH, MATH, U-MATH, MGSM, GPQA-Diamond, UGPhysics, KORBench), \emph{knowledge} (MMLU, MMLU-Redux, MMLU-Pro, C-Eval, SuperGPQA, NaturalQuestions, AGIEval-EN, INCLUDE-Base-44, SimpleQA, FLORES, Belebele), \emph{reasoning and mathematical word-problem solving} (HellaSwag, SIQA, DROP, BBH, GSM8K), and \emph{code} (HumanEval+, MBPP+, LiveCodeBench, BigCodeBench, BigCodeBench-Hard). Benchmark sources are: STEM~\citep{wei2023cmath,hendrycks2021math,chernyshev2025umath,shi2023mgsm,rein2024gpqa,xu2025ugphysics,ma2025korbench}; knowledge and multilingual suites~\citep{hendrycks2021mmlu,gema2025mmluredux,wang2024mmlupro,huang2023ceval,pteam2025supergpqa,kwiatkowski2019naturalquestions,zhong2024agieval,romanou2025include,wei2024simpleqa,goyal2022flores,bandarkar2024belebele}; reasoning~\citep{zellers2019hellaswag,sap2019socialiqa,dua2019drop,suzgun2023bbh,cobbe2021gsm8k}; and code~\citep{chen2021codex,austin2021program,liu2023evalplus,jain2025livecodebench,zhuo2025bigcodebench}. We follow each benchmark's standard protocol in a unified evaluation harness: multiple-choice knowledge benchmarks use few-shot in-context likelihood (PPL) scoring, while generative benchmarks (math, code, open-ended QA, GPQA) use greedy decoding (temperature 0) unless a benchmark's standard recipe specifies sampling, in which case we report the mean over samples (e.g., GPQA-Diamond \texttt{avg@5}).


\section{Prompts}
\label{app:prompts}

For reproducibility, we reproduce the core prompts used by the pipeline. They are lightly edited for length and formatting; the operative instructions are verbatim. Query generation and TOC planning use DeepSeek-V3.2; the TOC quality gate is judged by DeepSeek-V3.2; section content is generated by Qwen3.5-35B-A3B.

\subsection{Query Generation (Statement Form)}
\label{app:prompt:query}

\begin{quote}\small\ttfamily
\textbf{[System]} You are a distinguished academic expert and curriculum architect. Your task is to design high-quality declarative retrieval statements for educational corpus search.

\textbf{[User]} \# Context\\
We are preparing retrieval inputs for the sub-discipline below: ``\{current\}'' (Classification: \{discipline\_path\}).

\# Task\\
Generate 10 to 15 retrieval queries in \textbf{declarative sentence style}.

\# Requirements\\
1. \textbf{Declarative style only}: every query must be a complete declarative sentence, not a question and not a pure keyword list.\\
2. \textbf{MECE coverage}: cover the full scope with minimal overlap (concepts, mechanisms, methods, applications, limitations, frontier).\\
3. \textbf{Retrieval-friendly}: concise but information-dense, using accurate domain terminology.\\
4. \textbf{Multi-granularity}: include both foundational and advanced statements to support broad and deep recall.\\
5. \textbf{Language}: write all queries in English.

\# Output Format\\
Respond strictly with a JSON object: \{``sub\_discipline'': ..., ``queries'': [\{``angle'': ..., ``search\_query'': ..., ``rationale'': ...\}, ...]\}.
\end{quote}

\subsection{TOC Planning}
\label{app:prompt:toc}

\begin{quote}\small\ttfamily
\textbf{[System]} You are an expert content architect. Your task is to study the discipline information and topic summaries provided by the user, think carefully, and then produce a structured, hierarchical table of contents. Adapt the structure naturally to the requested audience and writing style. Your final answer must contain only one \texttt{json ...} code block.

\textbf{[User]} provides the sub-discipline, its taxonomy path and scope
description, and the candidate topic summaries in the form
\texttt{[cluster\_id=id] theme}. Each request also includes:\\
\medskip
{[Variant]}\\
This is variant \{i\}/\{k\} for the same cluster set. It should be
structurally different from the other variants.\\
\medskip
{[Audience and Style]}\\
This variant is for \{target\_audience\} and should follow a
\{writing\_style\} style. Adjust depth, terminology density, and organization
accordingly.\\
{[Audience Requirements]}\\
\{audience-specific planning brief\}\\
{[Style Requirements]}\\
\{style-specific planning brief\}\\
\medskip
{[A. Structure and Fields]}\\
1. Use at most three hierarchy levels; every node has a title and description.\\
2. Non-leaf nodes contain only \texttt{title}, \texttt{description}, and
\texttt{children}. Leaf nodes contain only \texttt{title},
\texttt{description}, \texttt{clusters}, and \texttt{token\_count}.\\
3. Every leaf must cite at least one supporting cluster and an integer word
budget. A leaf without suitable evidence must be merged or removed.\\
4. Cite at most three clusters per leaf, selected for both topical fit and the
target audience. Prefer not to reuse one cluster more than three times.\\
\medskip
{[B. Coverage and Budget]}\\
5. The sum of leaf budgets should generally not exceed
\{max\_total\_words\}; narrower or weakly supported topics should be shorter.\\
6. Do not force every cluster into the TOC. Omit weak, incoherent, overly
advanced, or overly basic themes when this improves the book structure.\\
\medskip
{[C. Readability and Naming]}\\
7. Make the hierarchy clear and navigable, with consistent local numbering or
labels that expose parent--child relationships.\\
8. Do not assume images or figures.\\
\medskip
{[Output]}\\
Return one JSON tree. Top-level and intermediate nodes use
\texttt{\{title, description, children\}}; leaves use
\texttt{\{title, description, clusters, token\_count\}}.
\end{quote}

\noindent The audience/style briefs at this stage control the organization of
the book, not merely its eventual prose: for example, the college/textbook
profile requests a layered instructional path with room for derivations and
worked cases, whereas practitioner/tutorial profiles emphasize workflows,
decisions, constraints, and executable steps.

\subsection{TOC Quality Gate}
\label{app:prompt:tocjudge}

The judge receives the discipline, \texttt{target\_audience}, \texttt{writing\_style}, the total token budget, and the TOC tree (each leaf annotated with its referenced \texttt{cluster\_themes} and \texttt{token\_count}). It is instructed to evaluate \emph{against the target style rather than a single textbook standard} and to return a JSON object:

\begin{quote}\small\ttfamily
\{\\
\ \ ``score'': 1--5 (5 = excellent, 4 = continue, 3 = borderline, 2 = do not continue, 1 = severely unfit),\\
\ \ ``recommend\_continue'': bool (typically true only when score $\geq$ 4 and no mid/high-severity issue),\\
\ \ ``audience\_fit\_score'': 1--5,\ ``style\_fit\_score'': 1--5,\\
\ \ ``structure\_score'': 1--5,\ ``support\_score'': 1--5,\\
\ \ ``token\_allocation\_score'': 1--5,\ ``generation\_readiness\_score'': 1--5\\
\}
\end{quote}
A TOC proceeds to content generation only if \texttt{recommend\_continue} is true. The six per-dimension scores (audience fit, style fit, structure, support, token allocation, and generation readiness) are the criteria underlying that decision; Table~\MainPassRateTable{} reports the resulting pass rates by audience$\times$style rather than these raw dimension scores.

\subsection{Source-Grounded Section Generation}
\label{app:prompt:section}

\begin{quote}\small\ttfamily
\textbf{[System]} You are an expert textbook writer. You turn scattered reference materials into coherent, well-structured, high-density textbook section prose.

Writing requirements (abridged):\\
1. Use only the strongly relevant parts of the reference materials; discard noise and do not force in irrelevant material.\\
2. Do not merely paraphrase; synthesize and reorganize into a logically rigorous, step-by-step narrative fitting the requested audience and style.\\
3. Preserve/normalize formulas in LaTeX (\$...\$, \$\$...\$\$) and use Markdown tables/lists where helpful; avoid walls of plain text.\\
4. Be faithful to the materials; do \emph{not} invent specific numbers, years, names, papers, DOIs, theorem names, or experimental conclusions; correct obvious OCR/transcription errors or omit them.\\
5. Produce clean reader-facing prose: never mention ``reference material'', ``retrieved text'', snippet/source labels, or image-dependent phrasing; no HTML/XML tags.\\
6. Do not repeat the section title (added at assembly) and do not invent global chapter numbering that conflicts with the book.\\
7. Treat the requested length as a target for detail, not a hard lower bound; prioritize density and accuracy over padding.\\
8. Use the supplied TOC context to differentiate this section from adjacent sections, avoid repetition, and add natural transitions where helpful.\\
9. Write the final content in English.

\textbf{[User]} provides the following section identity and structural
context, followed by the filled audience/style profile and the $k{=}10$
sampled source chunks:\\
\medskip
Discipline path: \{taxonomy path\}\\
Section position: \{full title path from book root to current leaf\}\\
Section title: \{leaf title\}\\
Section overview: \{leaf description\}\\
\medskip
{[Structural Context (TOC Tree)]}\\
- Current section position: \{chapter/section index path\}\\
- Marker: ``$>>$'' indicates the current node.\\
{[Current Level (Sibling Sections)]}\\
\{preceding sibling titles and descriptions\}\\
$>>$ \{current leaf title and description\}\\
\{following sibling titles and descriptions\}\\
{[Parent Level and Its Siblings]}\\
\{parent title/description together with the preceding and following
parent-level nodes\}\\
Use this context to keep the section clearly differentiated from siblings,
avoid repetition, and add natural transitions where helpful.\\
\medskip
Target audience: \{target\_audience\}\\
Writing style: \{writing\_style\}\\
\medskip
{[Audience requirements]}\\
\{audience-specific brief\}\\
\medskip
{[Style requirements]}\\
\{style-specific brief\}\\
\medskip
Adjust depth, terminology density, and expression to the audience and style while preserving factual accuracy.
\end{quote}

\paragraph{Injected structural context.}
The generator therefore sees both the full title path and a local TOC
neighborhood: the current leaf's preceding and following siblings, the parent
node, and the parent's neighboring nodes, each with its description. This
context is used to preserve section boundaries, reduce cross-section
duplication, and support coherent transitions.

\paragraph{Injected audience/style briefs.}
The production prompt expands the two placeholders above rather than passing
only their labels. The audience briefs require: \texttt{high\_school\_students},
little assumed knowledge, intuitive explanations, thought experiments, and
real-life scenarios; \texttt{college\_students}, systematic and in-depth
development with supported examples, formulas, or derivations;
\texttt{practitioners}, application contexts, workflows, decision criteria,
implementation constraints, and risk boundaries; and \texttt{researchers},
academic high-density prose covering advanced theory, methodological
comparisons, evidence, controversies, and boundary conditions. The style briefs
separately require: \texttt{textbook}, systematic, rigorous, professional, and
applied exposition; \texttt{blog}, verifiable information in an approachable
voice with supported examples or anecdotes; \texttt{tutorial}, reusable
step-by-step explanation with motivations, caveats, and practical rules; and
\texttt{wiki}, neutral, standardized, retrieval-friendly organization that can
also function as a standalone reference entry.

\subsection{Intrinsic-Quality Judge Rubrics}
\label{app:prompt:intrinsic_judges}

The production judge prompts were written in Chinese; below we provide
condensed English versions of the reported evaluation criteria.
Gemini-3.1-Pro
receives the complete generated text together with its title, target audience,
and writing style, and returns only a JSON score object. The same model and
rubric are used for every condition within each ablation.

\paragraph{Section-level chunk-count rubric (1--5).}
The judge assesses whether a generated section is useful as a training sample.
The overall anchors are: 5, excellent---factually robust, coherent, educational,
and nearly ready for training; 4, good---minor issues but suitable for training;
3, useful but with clear weaknesses; 2, poor---issues likely to harm training;
and 1, unsuitable. It independently scores:
\begin{itemize}
\item \textbf{Factual accuracy}: reliability of facts, definitions, formulas,
and derivations, and absence of evident hallucinations.
\item \textbf{Theme alignment}: focus on the section title and intended topic.
\item \textbf{Structure/coherence}: progression, transitions, and repetition
control.
\item \textbf{Educational value}: clarity, conceptual progression, examples,
and summaries.
\item \textbf{Audience/style fit}: appropriate difficulty, terminology density,
and register.
\item \textbf{Breadth}: distinct relevant subtopics or perspectives rather than
repetition of one point.
\item \textbf{Depth}: mechanisms, principles, derivations, causal explanations,
boundary conditions, or counterexamples rather than definition-level coverage.
\end{itemize}
Breadth and depth are judged independently, with 3 as the ordinary acceptable
level; scores above 3 require identifiable evidence in the complete section.
The judge is instructed not to reward length, fluency, or formatting by
themselves, and not to infer factual errors without evidence. The prompt returns
the seven reported dimensions plus an overall score.

\paragraph{Book-level component rubric (1--10).}
The judge assesses the value of a complete book as mid-training annealing data,
using the full scale with anchors 5 (marginally acceptable), 7 (good), and
9--10 (excellent). It scores nine dimensions plus an overall score:
\begin{itemize}
\item \textbf{Factuality}: factual and terminological correctness.
\item \textbf{Specificity}: verifiable domain details, explicit conditions and
boundaries, and avoidance of generic expansion.
\item \textbf{Breadth}: systematic coverage and absence of major omissions.
\item \textbf{Depth}: mechanisms, principles, and derivations beyond definitions.
\item \textbf{Structure}: book hierarchy, cross-section progression, and
control of jumps, duplication, and outline-like fragmentation.
\item \textbf{Pedagogy}: clarity of definitions, explanations, and derivations.
\item \textbf{Originality}: useful organization or explanatory perspective
without sacrificing factuality.
\item \textbf{Example quality}: concrete, verifiable, topic-relevant examples.
\item \textbf{Audience/style fit}: matching difficulty, terminology, depth, and
register to the requested profile.
\end{itemize}
The judge is instructed to evaluate training value rather than surface
fluency or Markdown quality. Short, outline-like, generic, repetitive, or
off-topic text lowers the relevant dimensions and overall score; weak
specificity or depth prevents a high overall score.

\input{appendix_case_study.tex}

\section{Query-Form Pilot}
\label{app:query_form_pilot}

Before fixing the retrieval-query format, we compared statement-, question-, and keyword-style queries using the same type of downstream cluster relevance filter used in the production pipeline. For each condition, retrieved passages were clustered and an LLM judged whether each cluster should be included in the textbook for the target discipline. The production filter follows the same principle but uses a slightly more permissive prompt, so that the final corpus preserves broader source coverage when a cluster is still plausibly useful. We use the pilot retention rate as a proxy for whether the query form retrieves material that forms coherent, in-scope textbook subtopics. Statement-form queries had the highest cluster retention rate: 78.8\% of clusters were retained, compared with 61.9\% for question-style queries and 56.9\% for keyword-style queries.

Qualitatively, the lower retention rates for question and keyword queries match the failure modes we observed during inspection. Keyword queries are compact but often underspecified: a query such as a short list of terms can retrieve passages that share vocabulary while mixing several unrelated subtopics, yielding clusters that the relevance filter rejects as diffuse. Question-style queries are more specific, but their interrogative form often favors short answer-oriented or FAQ-like passages and can overemphasize a single requested formulation. Statement-form queries instead resemble the declarative exposition common in books, papers, and web documents; they tend to retrieve passages that state, elaborate, or contextualize the same proposition, making the resulting clusters more coherent and easier to map into textbook subtopics. This pilot motivated the use of declarative statement-form queries throughout the final pipeline.

\section{Random Concatenation Control}
\label{app:random_concat}

\noindent
\begin{minipage}[t]{0.48\linewidth}
\vspace{0pt}
To test whether the Full--Split gap is explained by longer documents or fewer
boundary resets alone, we evaluate a length-matched RandomConcat control that
randomly concatenates Split sections to match the Full document-length
distribution while destroying planned book coherence.
Table~\ref{tab:random_concat} reports category means. RandomConcat is nearly
identical to Split on the 28-benchmark mean (54.89 vs.\ 54.88) and remains below
Full (55.90; $+1.01$ Full--RandomConcat), suggesting that length matching alone
does not recover the benefit of coherent book-level organization.
\end{minipage}\hfill
\begin{minipage}[t]{0.49\linewidth}
\vspace{0pt}
\centering
\footnotesize
\captionsetup{type=table,position=top,width=\linewidth,font=small,skip=6pt,justification=raggedright,singlelinecheck=false}
\setlength{\tabcolsep}{2pt}
\caption{Length-matched random concatenation control.
$\Delta_{\mathrm{F}}=$ Full$-$RandomConcat and
$\Delta_{\mathrm{S}}=$ RandomConcat$-$Split.}
\label{tab:random_concat}
\begin{tabular}{@{}l c c c c c@{}}
\toprule
\textbf{Category} & \textbf{RandomConcat} & \textbf{Full} &
\textbf{Split} & \textbf{$\Delta_{\mathrm{F}}$} &
\textbf{$\Delta_{\mathrm{S}}$} \\
\midrule
STEM      & 47.19 & 47.86 & 47.30 & +0.67 & $-$0.11 \\
Knowledge & 54.89 & 55.04 & 54.43 & +0.15 & +0.46 \\
Reasoning & 77.86 & 78.78 & 77.43 & +0.92 & +0.43 \\
Code      & 42.69 & 46.18 & 43.95 & +3.49 & $-$1.26 \\
\midrule
Overall   & 54.89 & 55.90 & 54.88 & +1.01 & +0.01 \\
\bottomrule
\end{tabular}
\end{minipage}
\par\medskip

\section{Llama3-8B Transfer Experiment}
\label{app:llama_transfer}

We additionally test whether the training-data effect transfers to a different
architecture and scale using Llama3-8B. We compare Natural Books with Full and
RandomConcat. Each condition is trained on the same 100B-token mixture with an
8B-token book slice (8\%). Data order, training steps, optimizer, and
learning-rate schedule are held fixed; only the book condition changes. The
three runs require 39.9K H800 GPU-hours in total. Training uses 256 H800 GPUs, a global batch size of
512 (2 examples per GPU), and AdamW with learning rate $5{\times}10^{-5}$,
weight decay 0.1, $\beta_1{=}0.9$, $\beta_2{=}0.95$, and gradient clipping at
1.0. We train for 6{,}373 steps using cosine decay with a 10\% warmup and a
minimum learning-rate ratio of 0.1, bfloat16 precision, gradient checkpointing,
FSDP auto-wrapping, seed 42, and Transformers 4.44.2. The 28-benchmark transfer
suite spans four categories: STEM, knowledge and multilingual understanding,
reasoning and mathematical word-problem solving, and code. Category means are
reported in Table~\MainLlamaTable{} of the main paper.

\begin{wraptable}[9]{r}{0.49\linewidth}
\vspace{-9pt}
\centering
\footnotesize
\captionsetup{type=table,position=top,width=\linewidth,font=small,skip=6pt,justification=raggedright,singlelinecheck=false}
\setlength{\tabcolsep}{2pt}
\caption{Benchmark-level robustness for the Llama3-8B transfer experiment.
Intervals are 95\% CIs. Full$-$Natural has one tied benchmark, excluded from
its sign test.}
\label{tab:llama_robustness}
\begin{tabular}{@{}l c c c c@{}}
\toprule
\textbf{Comparison} & \textbf{Mean $\Delta$} & \textbf{95\% CI} &
\textbf{W/L} & \textbf{Sign $p$} \\
\midrule
Full $-$ Random  & +0.86 & [+0.44,\ +1.30] & 20/8 & 0.018 \\
Full $-$ Natural & +1.52 & [+0.55,\ +2.76] & 22/5 & 0.00076 \\
Random $-$ Natural & +0.66 & [$-$0.23,\ +1.75] & 17/11 & 0.172 \\
\bottomrule
\end{tabular}
\vspace{-2pt}
\end{wraptable}
Full improves the 28-benchmark mean by $+1.52$ over Natural Books and by
$+0.86$ over RandomConcat, with 20 wins and 8 losses. The advantage is positive
in all four categories (STEM $+1.49$, reasoning $+1.17$, code $+1.02$,
knowledge $+0.24$).

We assess benchmark-level robustness using the same paired procedure as in
Appendix~\SuppBootstrapAppendix{}: 100{,}000 bootstrap resamples of benchmark
identities and a one-sided sign test over benchmark wins. These checks support
Full over RandomConcat and Natural Books at the benchmark level. By contrast,
the RandomConcat--Natural interval crosses zero and its sign test is not
significant, so the paired evidence specifically favors Full rather than
arbitrary length-matched concatenation. These benchmark-level checks do not
estimate training-run variance. The same paired bootstrap and sign-test
protocol is reused for the main MoE suite in Appendix~\SuppBootstrapAppendix{}.
\WFclear

\section{Benchmark-Suite Robustness}
\label{app:benchmark_bootstrap}

\begin{wraptable}[9]{r}{0.49\linewidth}
\vspace{-11pt}
\centering
\footnotesize
\captionsetup{type=table,position=top,width=\linewidth,font=small,skip=6pt,justification=raggedright,singlelinecheck=false}
\setlength{\tabcolsep}{1.5pt}
\caption{Benchmark-level 95\% CIs and sign tests for the aggregate deltas in
Table~\MainBenchmarkTable{}. Sign-test $p$ values are one-sided.}
\label{tab:benchmark_bootstrap}
\begin{tabular}{@{}l c c c c@{}}
\toprule
\textbf{Comparison} & \textbf{Mean $\Delta$} &
\textbf{95\% CI} & \textbf{W/L} & \textbf{Sign $p$} \\
\midrule
Full $-$ Natural Books & +1.09 & [+0.41,\ +1.87] & 19/9 & 0.044 \\
Full $-$ Split         & +1.02 & [+0.45,\ +1.69] & 22/6 & 0.0019 \\
Full $-$ Rephrase      & +1.17 & [+0.70,\ +1.69] & 25/3 & 0.000014 \\
\bottomrule
\end{tabular}
\vspace{-2pt}
\end{wraptable}
The main table reports unweighted averages over a 28-benchmark suite. As
benchmark-level robustness checks for this aggregate reporting, we run a paired
bootstrap over benchmarks and a sign test over benchmark wins. Each bootstrap
replicate resamples the 28 benchmark identities with replacement, using the
same sampled benchmark set for Natural Books, Split, Rephrase, and Full; it then
recomputes the unweighted mean delta across benchmarks. We repeat this procedure
100{,}000 times and report percentile 95\% intervals. The sign test counts how
many benchmarks Full exceeds each comparator and applies a one-sided binomial
test under a 50/50 null. All three intervals exclude zero; the Rephrase
comparison is strongest by both win count (25/3) and interval lower bound
($+0.70$).
\WFclear

\section{Component Ablation Details}
\label{app:component_ablation}

Table~\ref{tab:retrieval_ablation} reports all dimensions from the component
ablation summarized in \S\MainComponentSection{}. The rubric follows the
book-level judge in Appendix~\ref{app:prompt:intrinsic_judges}: factuality checks
correctness, while specificity rewards verifiable domain details and explicit
conditions. Breadth measures systematic coverage, whereas depth requires
mechanisms, principles, or derivations beyond definitions. Structure evaluates
hierarchy, cross-section progression, and repetition control; pedagogy evaluates
the clarity of definitions, explanations, and derivations. Originality captures
useful organization or explanatory perspective, examples must be concrete and
topic-relevant, and audience/style fit checks difficulty, terminology, depth,
and register.

\smallskip
Full leads on overall quality, factuality, specificity, depth, originality, and
examples. Baseline A remains strongest on breadth, structure, pedagogy, and
audience/style fit because it retains Full's cluster-informed TOC. Baseline C's
large drops in depth, examples, and originality diagnose the cost of direct
whole-book generation, while Baseline D shows the smaller but broad degradation
from using a rejected TOC.

\begin{table}[!htbp]
\centering
\footnotesize
\captionsetup{position=top,width=\linewidth,font=small,skip=6pt,justification=raggedright,singlelinecheck=false}
\setlength{\tabcolsep}{3pt}
\caption{Component ablation results (1--10 scale). Full is the complete
pipeline; Baseline A removes source chunks, Baseline B also removes
cluster-informed TOC planning, Baseline C directly generates a full book, and
Baseline D uses a failed TOC.}
\label{tab:retrieval_ablation}
\begin{tabular}{@{}l cccccccccc@{}}
\toprule
\textbf{Condition} & \textbf{Overall} & \textbf{Factual} &
\textbf{Spec.} & \textbf{Breadth} & \textbf{Depth} & \textbf{Struct.} &
\textbf{Ped.} & \textbf{Orig.} & \textbf{Ex.} & \textbf{Aud.\ fit} \\
\midrule
Full Pipeline & \textbf{9.38} & \textbf{9.52} & \textbf{9.52} & 9.45 &
\textbf{9.43} & 9.28 & 9.14 & \textbf{8.76} & \textbf{9.24} & 9.12 \\
\midrule
Baseline A & 9.21 & 9.20 & 9.29 & \textbf{9.47} & 9.34 &
\textbf{9.35} & \textbf{9.23} & 8.67 & 8.89 & \textbf{9.21} \\
Baseline B & 9.07 & 9.24 & 9.33 & 9.44 & 9.27 &
8.91 & 9.00 & 8.59 & 8.86 & 8.62 \\
Baseline C & 6.12 & 7.73 & 6.41 & 8.15 & 5.86 &
6.42 & 6.92 & 5.58 & 5.46 & 6.25 \\
Baseline D & 9.13 & 9.34 & 9.36 & 9.42 & 9.28 &
9.08 & 8.95 & 8.57 & 9.02 & 8.75 \\
\bottomrule
\end{tabular}
\end{table}

\section{Decontamination Details}
\label{app:decontam}

\paragraph{Procedure.}
We decontaminate against benchmark test sets using 13-gram overlap over
a mixed Chinese/English tokenization: English is split into words (1-gram per
token) and each Chinese character counts as 0.5-gram. The same 13-gram
representation is computed for every benchmark item in the evaluation resources.
For final generated books, we concatenate the \texttt{title} and \texttt{text}
fields (for QA-style records, \texttt{input}+\texttt{output}) and match them
against the union of test-set n-grams. The filter is deliberately conservative:
a document is labeled contaminated if it shares \emph{any} 13-gram with a
benchmark test item (match-fraction threshold of $0$), so even a single
overlapping 13-gram triggers a flag.
Only the \texttt{valid/} split is used for training. We also run the same
filter on the retrieved source-document pool used by the generator as a
source-level audit. Each flagged record stores an \texttt{extra\_info} field
(matched n-grams, match/no-match ratios, source benchmark, and contamination
level) for inspection.

\paragraph{Result.}
Tables~\ref{tab:decontam_source} and~\ref{tab:decontam_generated} report the
portion of the decontamination results that overlaps the benchmarks in
Table~\MainBenchmarkTable{}. The source share uses the full pool (85B valid
$+$ 0.95B flagged across all evaluation sources); the 60.2M tokens shown are
the Table~\MainBenchmarkTable{}-overlapping portion. In the retrieved
source-document pool, the filter flags 12{,}825 source records, corresponding to 60.2M tokens
($\sim$0.07\% of the source-document token pool). In the final generated corpus,
it removes 579 generated books, corresponding to 22.9M tokens
($\sim$0.07\% of the generated corpus). The check captures \emph{verbatim}
13-gram overlap only and does not rule out paraphrased leakage; because the
matcher flags even partial overlaps, the counts are upper bounds on true
test-item leakage.

\par\smallskip\noindent
\begin{minipage}[t]{0.49\linewidth}
\centering
\footnotesize
\renewcommand{\arraystretch}{1.05}
\captionsetup{type=table,position=top,width=\linewidth,font=small,skip=4pt,justification=raggedright,singlelinecheck=false}
\setlength{\tabcolsep}{3pt}
\caption{Retrieved source documents flagged by overlap with the reported
benchmarks; share is relative to the full source-document token pool.}
\label{tab:decontam_source}
\begin{tabular}{@{}l r r r@{}}
\toprule
\textbf{Benchmark} & \textbf{Records} & \textbf{Tokens} & \textbf{Share} \\
\midrule
SuperGPQA       & 5{,}320 & 25.2M & 0.029\% \\
MMLU-Pro        & 2{,}867 & 13.7M & 0.016\% \\
UGPhysics       & 2{,}538 & 12.5M & 0.015\% \\
MMLU-Redux      & 1{,}770 &  7.1M & 0.008\% \\
MATH            &   208  &  1.1M & 0.001\% \\
U-MATH          &    80  &  0.5M & 0.001\% \\
BBH             &    41  &  0.2M & $<$0.001\% \\
GPQA-Diamond    &     1  & $<$0.1M & $<$0.001\% \\
\midrule
Total  & 12{,}825 & 60.2M & 0.070\% \\
\bottomrule
\end{tabular}
\end{minipage}\hfill
\begin{minipage}[t]{0.46\linewidth}
\centering
\footnotesize
\renewcommand{\arraystretch}{1.05}
\captionsetup{type=table,position=top,width=\linewidth,font=small,skip=4pt,justification=raggedright,singlelinecheck=false}
\setlength{\tabcolsep}{3pt}
\caption{Generated books removed due to overlap with the reported benchmarks;
share is relative to the 32B-token generated corpus.}
\label{tab:decontam_generated}
\begin{tabular}{@{}l r r r@{}}
\toprule
\textbf{Benchmark} & \textbf{Books} & \textbf{Tokens} & \textbf{Share} \\
\midrule
UGPhysics       & 184 & 7.7M & 0.025\% \\
SuperGPQA       & 179 & 6.4M & 0.021\% \\
MMLU-Pro        & 147 & 6.2M & 0.020\% \\
MATH            &  31 & 1.4M & 0.004\% \\
MMLU-Redux      &  36 & 1.1M & 0.004\% \\
BBH             &   2 & 0.1M & $<$0.001\% \\
\midrule
Total  & 579 & 22.9M & 0.074\% \\
\bottomrule
\end{tabular}
\end{minipage}
\par\smallskip
\paragraph{Code availability.}
We submit, as supplementary material, a reference implementation of the
retrieval-grounded textbook synthesis pipeline and of the procedures used to
construct the Split, RandomConcat, and Rephrase control sets, together with
paper-aligned configuration templates. The package is provider-agnostic: chat,
embedding, and retrieval backends are configured
externally, and a bundled smoke-test configuration runs without network access.
It does not include the proprietary retrieval corpus or index, the Natural Books
mixture slice, model checkpoints, mid-training code, or the downstream
evaluation harness; reproducing the reported scores therefore requires
equivalent user-provided data and compute. Stochastic provider outputs can also
differ from the production run even under the same logical configuration.

\paragraph{Data availability.}
The in-house retrieval corpus and original natural-book slice cannot be
redistributed in full because of source licensing constraints. Public
substitutes would alter both the source pool used for the retrieval-pool-matched
comparison and the original mid-training mixture used by Natural Books, undermining
the retrieval-pool, content, and recipe controls central to this study. We
will release a research-licensed subset of the synthetic textbook corpus upon
publication.

%% file: appendix_case_study.tex
\section{End-to-End Running Example}
\label{app:running_example}

This appendix traces one real pipeline instance for the discipline
\emph{Feedforward Neural Networks and Backpropagation} (ID 520206001), the
running example in \S\MainMethodSection{}. We show the output of each stage
separately, followed by representative generated-section excerpts. Apart from
the explicitly marked omissions and Markdown-to-\LaTeX{} typesetting, the
displayed content is taken directly from the production artifacts.

\subsection{Step 0: Discipline and Query Generation}

\paragraph{Taxonomy input.}
\emph{Computer Science and Technology $\rightarrow$ Artificial Intelligence
$\rightarrow$ Computer Neural Networks $\rightarrow$ Feedforward Neural
Networks and Backpropagation}. The taxonomy description specifies the
architecture and training of multilayer perceptrons, including the mathematical
derivation of error backpropagation.

\paragraph{Generated retrieval statements.}
The complete set contains 14 statements:
\begin{enumerate}
\item The multilayer perceptron is a foundational type of feedforward neural network architecture.
\item Universal approximation theorems provide the theoretical justification for the representational power of feedforward neural networks.
\item The forward pass in a feedforward network computes activations layer by layer from input to output.
\item The backpropagation algorithm efficiently computes gradients of a loss function with respect to all network parameters.
\item The chain rule of calculus forms the mathematical foundation for deriving the backpropagation algorithm.
\item Stochastic gradient descent is commonly used to update weights during the training of multilayer perceptrons.
\item Vanishing and exploding gradients are significant challenges in training deep feedforward networks.
\item Activation functions introduce non-linearity into neural networks, enabling them to learn complex mappings.
\item Feedforward neural networks serve as the backbone for many deep learning models in computer vision and natural language processing.
\item Standard feedforward networks are stateless and cannot inherently model sequential dependencies.
\item Weight initialization strategies, such as He or Xavier initialization, are critical for effective training of deep feedforward networks.
\item Research on feedforward architectures explores alternatives to backpropagation, such as synthetic gradients or layer-wise training.
\item The computational graph is a useful abstraction for visualizing and implementing the forward and backward passes in neural networks.
\item The loss function quantifies the discrepancy between the network's predictions and the true target values.
\end{enumerate}

\subsection{Step 1: Retrieval}

The 14 statements produced 1{,}400 retrieval results and 1{,}346 unique
documents after deduplication. For example, the universal-approximation query
retrieved the following source:

\begin{quote}
\emph{Query.} Universal approximation theorems provide the theoretical
justification for the representational power of feedforward neural networks.

\emph{Retrieved passage (excerpt).} ``In this work we take a look at the
universal approximation question for stochastic feedforward neural networks.
In contrast to deterministic neural networks, which represent mappings from a
set of inputs to a set of outputs, stochastic neural networks represent
mappings from a set of inputs to a set of probability distributions over the
set of outputs. \ldots{} We study the representational power of deep sigmoid
belief networks in terms of compositions of linear transformations of
probability distributions.''
\end{quote}

This is one of many retrieved documents; no single passage is used as the book
outline. The next stage reorganizes the full retrieval pool into semantic
units.

\subsection{Step 2: Semantic Clustering}

The 1{,}346 documents yield 6{,}713 chunks, which are grouped into 40 clusters.
The relevance filter retains 35 clusters and rejects five off-scope or
incoherent clusters. Representative retained clusters include:
\begin{itemize}
\item \texttt{C35} (176 documents, 217 chunks): feedforward network structure,
including input, hidden, and output layers and unidirectional information flow.
\item \texttt{C5} (93 documents, 170 chunks): universal approximation theory
and the capacity to approximate continuous functions in normed spaces.
\item \texttt{C9} (71 documents, 112 chunks): efficient gradient computation
using the chain rule, computational graphs, and automatic differentiation.
\item \texttt{C32} (120 documents, 211 chunks): Xavier, He, and related
initialization methods for stable backpropagation and convergence.
\item \texttt{C3} (89 documents, 129 chunks): physics-informed feedforward
networks for physical-system modeling.
\end{itemize}

\paragraph{Examples of rejected noise.}
C0 primarily concerns recurrent networks, sequence modeling, and translation;
C12 mixes symbolic logic, multi-agent systems, linguistic semantics, and neural
networks without a coherent feedforward-network focus. Both are excluded before
TOC planning.

\subsection{Step 3: TOC Planning and Quality Gate}

The pipeline generates 16 audience$\times$style plans for this discipline; 15
pass the quality gate. Below is the \emph{complete} college/textbook hierarchy,
including every chapter and section. Parentheses report the requested section
budget and supporting cluster IDs; the 16 sections allocate 56{,}600 tokens and
use 26 of the 35 retained clusters.

\begin{enumerate}
\item \emph{Part I: Architectural Foundations and Representational Power}
  \begin{itemize}
  \item 1.1 The Feedforward Architecture (3{,}800; C35, C31, C21)
  \item 1.2 Activation Functions: Properties and Impacts (3{,}500; C21, C28)
  \item 1.3 Universal Approximation Theorems (4{,}200; C5, C6, C14)
  \item 1.4 Depth, Width, and Expressivity Trade-offs (3{,}000; C14, C23)
  \end{itemize}
\item \emph{Part II: The Backpropagation Algorithm and Its Mechanics}
  \begin{itemize}
  \item 2.1 Loss Functions and the Optimization Objective (3{,}200; C30, C4)
  \item 2.2 Gradient Descent: The Foundational Optimizer (3{,}400; C2, C4)
  \item 2.3 Derivation of the Backpropagation Algorithm (4{,}500; C10, C9, C8)
  \item 2.4 Computational Graphs and Automatic Differentiation (3{,}600; C8, C9)
  \end{itemize}
\item \emph{Part III: Practical Training and Advanced Dynamics}
  \begin{itemize}
  \item 3.1 Weight Initialization Strategies (3{,}300; C32, C7)
  \item 3.2 Advanced Optimization Algorithms (3{,}800; C4, C2)
  \item 3.3 Vanishing and Exploding Gradients (3{,}200; C7, C28)
  \item 3.4 Hyperparameter Tuning and Experimental Methodology (3{,}500; C1, C36)
  \end{itemize}
\item \emph{Part IV: Advanced Architectures and Extensions}
  \begin{itemize}
  \item 4.1 Convolutional Neural Networks (CNNs) (4{,}000; C16, C26)
  \item 4.2 Backpropagation in Sequential Models (3{,}800; C18, C20, C29)
  \item 4.3 Physics-Informed and Scientific Machine Learning (3{,}000; C3)
  \item 4.4 Frontiers and Alternatives to Backpropagation (2{,}800; C22, C33)
  \end{itemize}
\end{enumerate}

\paragraph{Quality-gate decision.}
The judge assigns an overall score of 4/5 and recommends continuation.
Audience fit, style fit, token allocation, and generation readiness receive
5/5; structure and source support receive 4/5. It identifies two low-severity
limitations: Section 4.2 slightly broadens the scope from feedforward to
recurrent models, and Section 4.3 relies on only one source cluster. Retaining
these diagnostics here shows that the example is an actual accepted output,
not a retrospectively polished plan.

\subsection{Step 4: Source-Grounded Generation and Assembly}

The 15 accepted plans each produce three independently sampled books, yielding
45 books, 630 sections, and 10.5M characters for this discipline. For the
college/textbook plan above, the three sampled editions contain 257{,}466,
243{,}241, and 254{,}144 characters, respectively; each follows the same
four-part, 16-section plan.

We show four \emph{representative} section excerpts covering different roles in
the book: architectural foundations, the learning objective, backpropagation,
and experimental methodology. The excerpts come from independently sampled
editions of the same TOC. For space, we reproduce selected portions rather than
the complete sections. The text is verbatim apart from Markdown-to-\LaTeX{}
typesetting, and every omission is marked explicitly.

\input{appendix_case_selected_sections.tex}

%% file: appendix_case_selected_sections.tex

\subsubsection*{Section 1.2: Activation Functions: Properties and Impacts (Edition A)}
\begingroup
Activation functions constitute a fundamental component of the feedforward neural network architecture, serving as the mechanism that introduces non-linearity into the model. Without an activation function, a neural network would simply be a stack of linear transformations, equivalent to a single linear model regardless of its depth. This would severely limit the network's capacity to approximate complex, non-linear relationships present in real-world data. The selection of an appropriate activation function influences not only the representational power of the network but also the stability and efficiency of the training process, particularly regarding gradient flow during backpropagation.

\paragraph{Mathematical Role in Backpropagation}

In the context of a feedforward network, the computation proceeds layer by layer. For a neuron $j$ in layer $l$, the pre-activation value $z_j^{(l)}$ is computed as a weighted sum of the outputs from the previous layer plus a bias term:
\[
z_j^{(l)} = \sum_{i} w_{ji}^{(l)} a_i^{(l-1)} + b_j^{(l)}
\]
where $a_i^{(l-1)}$ is the activation from the preceding layer, $w_{ji}^{(l)}$ represents the weights, and $b_j^{(l)}$ is the bias. The activation $a_j^{(l)}$ is then obtained by applying the activation function $f(\cdot)$ to this sum:
\[
a_j^{(l)} = f(z_j^{(l)})
\]

During the training phase, the network minimizes an error function $E$ by updating weights using gradient descent. The Backpropagation algorithm computes the gradient of the error with respect to each weight $w_{ji}^{(l)}$ using the chain rule. A critical term in this derivation is the partial derivative of the error with respect to the pre-activation value, often denoted as the error signal $\delta_j^{(l)}$. This signal is defined as:
\[
\delta_j^{(l)} = \frac{\partial E}{\partial z_j^{(l)}} = \frac{\partial E}{\partial a_j^{(l)}} f'(z_j^{(l)})
\]
Here, $f'(z_j^{(l)})$ is the derivative of the activation function. The update rule for the weights then becomes proportional to this error signal multiplied by the input from the previous layer:
\[
\frac{\partial E}{\partial w_{ji}^{(l)}} = \delta_j^{(l)} a_i^{(l-1)}
\]
Consequently, the properties of $f'(z)$ directly dictate how gradient information propagates backward through the network. If the derivative is small or vanishes as the network depth increases, the weights in the earlier layers receive negligible updates, leading to slow or stalled learning. Conversely, a derivative that preserves signal magnitude facilitates more efficient convergence.
\endgroup

\subsubsection*{Section 2.1: Loss Functions and the Optimization Objective (Edition B)}
\begingroup
The fundamental objective of training a neural network is framed as an optimization problem. In this context, the learning process is defined by the task of finding a set of parameters—weights and biases, collectively denoted as $\boldsymbol{\theta}$—that best map input data to target outputs. Mathematically, given a dataset containing input-output pairs $(\boldsymbol{x}, \boldsymbol{y})$, the goal is to minimize a specific metric that quantifies the discrepancy between the network's prediction $\hat{\boldsymbol{y}} = f(\boldsymbol{x}; \boldsymbol{\theta})$ and the true label $\boldsymbol{y}$. This discrepancy is formalized through a loss function, often referred to as a cost function or error function. Consequently, the problem of learning can be expressed as finding the parameter vector $\boldsymbol{\theta}^*$ that minimizes the expected loss over the data distribution:

\[
\boldsymbol{\theta}^* = \arg\min_{\boldsymbol{\theta}} \mathbb{E}_{\boldsymbol{x}, \boldsymbol{y}} \left[ \mathcal{L}\left(f(\boldsymbol{x}; \boldsymbol{\theta}), \boldsymbol{y}\right) \right]
\]

In practice, since the true distribution is unknown, this objective is approximated using the empirical risk over the available training set. For a dataset of size $n$, the optimization target becomes a sum of individual errors, often structured as:

\[
\min_{\boldsymbol{\theta}} \mathcal{L}_{total}(\boldsymbol{\theta}) = \sum_{i=1}^{n} \ell\left(f(\boldsymbol{x}_i; \boldsymbol{\theta}), y_i\right) + R(\boldsymbol{\theta})
\]

Here, $\ell(\cdot, \cdot)$ denotes the loss incurred on a single sample, while $R(\boldsymbol{\theta})$ represents a regularization term designed to penalize overly complex models and prevent overfitting. The choice of $\ell(\cdot, \cdot)$ is not arbitrary; it is dictated by the nature of the task (regression vs. classification) and the statistical assumptions made about the noise in the data.

\paragraph{Statistical Foundations and Loss Selection}

The selection of a specific loss function is rooted in statistical theory. If the relationship between inputs and outputs is governed by additive Gaussian noise, the most natural choice is the Mean Squared Error (MSE). Conversely, if the output represents a categorical distribution, the goal is typically to maximize the likelihood of the observed data, which leads directly to the Cross-Entropy loss. Understanding these derivations provides insight into why certain loss functions are preferred for specific architectures and why others lead to poor convergence behavior.

\paragraph{Regression and Mean Squared Error}

In regression problems, the target $y$ is a continuous scalar or vector. The standard objective is to minimize the squared Euclidean distance between the predicted value and the observed value. For a single training example, the Mean Squared Error is defined as:

\[
\ell_{MSE}(\hat{y}, y) = \frac{1}{2}(y - \hat{y})^2
\]

The factor of $\frac{1}{2}$ is often included for convenience, as it cancels out when taking the derivative. For a batch of $N$ samples, the aggregate loss becomes:

\[
\mathcal{L}_{MSE}(\boldsymbol{\theta}) = \frac{1}{2N} \sum_{i=1}^{N} (y_i - f(\boldsymbol{x}_i; \boldsymbol{\theta}))^2
\]

This formulation has clear statistical justification. If we assume the target values are generated by the true function plus noise distributed according to a Gaussian (Normal) distribution with zero mean, maximizing the likelihood of the data is mathematically equivalent to minimizing the squared error. Specifically, under the assumption that $y = f(\boldsymbol{x}) + \epsilon$ where $\epsilon \sim \mathcal{N}(0, \sigma^2)$, the Maximum Likelihood Estimate (MLE) for the parameters $\boldsymbol{\theta}$ corresponds to the minimization of the sum of squared residuals.

\medskip\noindent\emph{[Intermediate paragraphs omitted for length.]}

\paragraph{Classification and Cross-Entropy}

\medskip\noindent\emph{[Intermediate paragraphs omitted for length.]}

The Cross-Entropy loss is derived from the principle of Maximum Likelihood for categorical variables. For a single sample where the true class is represented by a one-hot vector $\boldsymbol{y}$ and the network predicts probabilities $\boldsymbol{\hat{y}}$, the cross-entropy loss is defined as:

\[
\ell_{CE}(\boldsymbol{\hat{y}}, \boldsymbol{y}) = - \sum_{c} y_c \ln(\hat{y}_c)
\]

Since $\boldsymbol{y}$ is a one-hot vector (one element is 1, others are 0), this simplifies to the negative logarithm of the predicted probability assigned to the true class:

\[
\ell_{CE}(\boldsymbol{\hat{y}}, \boldsymbol{y}) = - \ln(\hat{y}_{true})
\]

For a batch of data, the empirical cross-entropy loss is the average over all samples:

\[
\mathcal{L}_{CE}(\boldsymbol{\theta}) = - \frac{1}{N} \sum_{i=1}^{N} \ln \left( f(\boldsymbol{x}_i; \boldsymbol{\theta})_{y_i} \right)
\]
\endgroup

\subsubsection*{Section 2.3: Derivation of the Backpropagation Algorithm (Edition B)}
\begingroup
\paragraph{The Gradient Problem in Neural Network Training}

Training a neural network fundamentally involves minimizing a loss function that measures the discrepancy between the network's predictions and the ground truth labels. As established in the preceding sections, stochastic gradient descent (SGD) and its variants rely on computing the gradient of the loss function with respect to every learnable parameter in the network. Let $\theta$ denote the collection of all parameters, including layer weights, bias terms, and potentially embedding matrices. The objective is to solve the optimization problem:
\[
\min_{\theta} \mathcal{L}(\theta)
\]
where $\mathcal{L}(\theta)$ is the loss. To perform a parameter update using gradient descent, one requires the gradient vector $\nabla_{\theta} \mathcal{L}$.

In a neural network with $L$ layers, the number of parameters can be substantial, often reaching into the millions. Calculating the gradient numerically, for instance by perturbing each parameter slightly, would require a separate forward pass for every single parameter. This approach is computationally prohibitive. The backpropagation algorithm provides the solution to this problem. It is a specific instance of reverse-mode automatic differentiation that computes the exact gradients in a number of steps proportional to the number of operations in the forward pass, typically requiring only two passes through the network data flow (one forward, one backward).

Technically, backpropagation is an efficient application of the chain rule of calculus. It decomposes the complex derivative of a high-dimensional function into a sequence of simpler derivatives, caching intermediary results (activations) computed during the forward pass to avoid redundant calculation. This section details the algebraic derivation of the algorithm for a standard feedforward network and then reinterprets this process through the lens of computational graphs.

\paragraph{Mathematical Preliminaries and Notation}

To derive the gradient updates systematically, we first establish a consistent notation for the network architecture. Consider a multilayer perceptron (MLP) with $L$ layers. The network processes an input vector $\mathbf{x}$, which is treated as the activation of layer $0$, denoted as $\mathbf{a}^{(0)} = \mathbf{x}$. The network computes a sequence of values through the layers, indexed by $l = 1, \dots, L$.

For any given layer $l$, the computation proceeds in two stages. First, a linear transformation is applied using a weight matrix $\mathbf{W}^{(l)}$ and a bias vector $\mathbf{b}^{(l)}$. This produces the linear pre-activation signal, denoted as $\mathbf{z}^{(l)}$. Second, a non-linear activation function $\phi(\cdot)$ is applied element-wise to produce the output of the layer, $\mathbf{a}^{(l)}$.

\[
\mathbf{z}^{(l)} = \mathbf{W}^{(l)} \mathbf{a}^{(l-1)} + \mathbf{b}^{(l)}
\]
\[
\mathbf{a}^{(l)} = \phi(\mathbf{z}^{(l)})
\]

The parameters of the network are collected in the set $\Theta = \{ \mathbf{W}^{(1)}, \mathbf{b}^{(1)}, \dots, \mathbf{W}^{(L)}, \mathbf{b}^{(L)} \}$. The input $\mathbf{a}^{(0)}$ and the true labels are fixed constants during the forward pass. The final output of the network, $\mathbf{a}^{(L)}$, is fed into a loss function $\mathcal{L}$. Common examples include the Mean Squared Error for regression or Cross-Entropy for classification.

In matrix calculus, we assume standard conventions where column vectors are used, and $\mathbf{M}^\top$ denotes the transpose. The derivative of a scalar loss with respect to a vector $\mathbf{v}$ results in a vector of the same dimension. Similarly, the derivative of a scalar with respect to a matrix results in a matrix of the same shape as the derivative variable.

\paragraph{The Chain Rule and Error Propagation}

The core mechanism of backpropagation is the chain rule. If a variable $u$ depends on $v$, and $v$ depends on $w$, the chain rule states that $\frac{\partial u}{\partial w} = \frac{\partial u}{\partial v} \frac{\partial v}{\partial w}$. In a neural network, the loss $\mathcal{L}$ depends on the parameters $\mathbf{W}^{(l)}$ through a complex chain of dependencies across all layers. We must compute:
\[
\frac{\partial \mathcal{L}}{\partial \mathbf{W}^{(l)}} \quad \text{and} \quad \frac{\partial \mathcal{L}}{\partial \mathbf{b}^{(l)}}
\]
for all $l$.

Directly applying the chain rule to the full expression is difficult. Instead, we introduce an auxiliary variable known as the \emph{error term} or \emph{local gradient}, denoted by $\boldsymbol{\delta}^{(l)}$. This term represents the gradient of the loss with respect to the pre-activation values $\mathbf{z}^{(l)}$:
\[
\boldsymbol{\delta}^{(l)} = \frac{\partial \mathcal{L}}{\partial \mathbf{z}^{(l)}}
\]
Computing $\boldsymbol{\delta}^{(l)}$ allows us to express the gradients of the weights and biases in a much simpler form, which we will derive below.

\medskip\noindent\emph{[Intermediate paragraphs omitted for length.]}

\paragraph{The Hidden Layer Gradient}

For hidden layers $l < L$, the loss does not depend on $\mathbf{z}^{(l)}$ directly, but rather through the pre-activations of the next layer, $\mathbf{z}^{(l+1)}$. We can expand the derivative using the chain rule again:
\[
\boldsymbol{\delta}^{(l)} = \frac{\partial \mathcal{L}}{\partial \mathbf{z}^{(l)}} = \frac{\partial \mathcal{L}}{\partial \mathbf{z}^{(l+1)}} \frac{\partial \mathbf{z}^{(l+1)}}{\partial \mathbf{z}^{(l)}}
\]
Substituting the definition of the error term for layer $l+1$, this becomes:
\[
\boldsymbol{\delta}^{(l)} = \left( \frac{\partial \mathcal{L}}{\partial \mathbf{z}^{(l+1)}} \right) \frac{\partial \mathbf{z}^{(l+1)}}{\partial \mathbf{z}^{(l)}}
\]
To compute the term $\frac{\partial \mathbf{z}^{(l+1)}}{\partial \mathbf{z}^{(l)}}$, we recall the forward pass equations:
\[
\mathbf{z}^{(l+1)} = \mathbf{W}^{(l+1)} \mathbf{a}^{(l)} + \mathbf{b}^{(l+1)}
\]
\[
\mathbf{a}^{(l)} = \phi(\mathbf{z}^{(l)})
\]
Combining these, we see that $\mathbf{z}^{(l+1)}$ depends on $\mathbf{a}^{(l)}$ through the linear transformation $\mathbf{W}^{(l+1)}$. Thus, the gradient propagates back from layer $l+1$ through the weights of that layer and the derivative of the activation function at layer $l$. The explicit recursive formula is:
\[
\boldsymbol{\delta}^{(l)} = \left( (\mathbf{W}^{(l+1)})^\top \boldsymbol{\delta}^{(l+1)} \right) \odot \phi'(\mathbf{z}^{(l)})
\]
This recursive relationship is the heart of backpropagation. It allows the error term of the previous layer $\boldsymbol{\delta}^{(l)}$ to be computed using the error term of the next layer $\boldsymbol{\delta}^{(l+1)}$, the weight matrix of the next layer $\mathbf{W}^{(l+1)}$, and the derivative of the activation at the current layer. By starting with $\boldsymbol{\delta}^{(L)}$ calculated from the loss and moving backward layer by layer to $l=1$, we can compute all error terms efficiently.

\paragraph{Weight and Bias Gradients}

Once the error terms $\boldsymbol{\delta}^{(l)}$ are computed for all layers, the gradients with respect to the parameters are straightforward. For the weights $\mathbf{W}^{(l)}$, the chain rule yields:
\[
\frac{\partial \mathcal{L}}{\partial \mathbf{W}^{(l)}} = \frac{\partial \mathcal{L}}{\partial \mathbf{z}^{(l)}} \frac{\partial \mathbf{z}^{(l)}}{\partial \mathbf{W}^{(l)}}
\]
Substituting $\boldsymbol{\delta}^{(l)} = \frac{\partial \mathcal{L}}{\partial \mathbf{z}^{(l)}}$ and the definition of $\mathbf{z}^{(l)}$, we obtain the outer product of the error signal and the input activations:
\[
\frac{\partial \mathcal{L}}{\partial \mathbf{W}^{(l)}} = \boldsymbol{\delta}^{(l)} (\mathbf{a}^{(l-1)})^\top
\]

\medskip\noindent\emph{[Intermediate paragraphs omitted for length.]}

Similarly, for the bias term $\mathbf{b}^{(l)}$, since $\mathbf{z}^{(l)} = \mathbf{W}^{(l)} \mathbf{a}^{(l-1)} + \mathbf{b}^{(l)}$, the derivative with respect to $\mathbf{b}^{(l)}$ is simply the error term itself:
\[
\frac{\partial \mathcal{L}}{\partial \mathbf{b}^{(l)}} = \boldsymbol{\delta}^{(l)}
\]
The bias update does not depend on the previous layer's activations, as bias shifts the activation threshold directly.
\endgroup

\subsubsection*{Section 3.4: Hyperparameter Tuning and Experimental Methodology (Edition A)}
\begingroup
\paragraph{Data Partitioning and Generalization}

A rigorous experimental methodology requires a clear separation between data used for training, validating, and testing. This separation is fundamental to diagnosing the learning behavior of a neural network. The dataset is typically partitioned into three distinct subsets: the training set, the validation set, and the test set. The training set is used to update the model parameters via backpropagation. The validation set serves as an intermediary evaluation set that is not used for parameter updates. It is primarily used for hyperparameter tuning and model selection. For example, if a student wishes to decide whether to use a learning rate of $10^{-3}$ or $10^{-4}$, they would compare the loss on the validation set after a fixed number of epochs for each choice. The test set is reserved strictly for the final evaluation of the model and should not influence any training or tuning decisions.

The distinction between training loss and validation loss is the primary indicator of overfitting. Overfitting occurs when the model achieves a very low training loss but fails to generalize to unseen data, resulting in a significantly higher validation loss. This phenomenon indicates that the model has "learned by memory," capturing noise and fluctuations in the training data rather than the underlying patterns. Mathematically, we are interested in the generalization error, which measures the expected loss on the true data distribution, rather than the empirical error calculated on the finite training set. A low training loss is a necessary condition for a good model, but it is not sufficient. If the gap between training and validation loss widens significantly as training proceeds, it suggests that the model's capacity is exceeding the information content of the dataset, or that the regularization parameters need adjustment.

Monitoring these metrics allows for the implementation of early stopping strategies. Early stopping involves monitoring the validation loss during training and halting the process when the loss fails to improve for a specified number of epochs. This prevents the model from overfitting the training data and often yields a model that performs better on the test set than one trained for the maximum number of epochs. This strategy relies entirely on the existence of a hold-out validation set, emphasizing the necessity of proper data partitioning in any experimental design.

\paragraph{Systematic Tuning Strategies}

Given the sensitivity of neural networks to hyperparameters, systematic search strategies are required to identify effective configurations. A naive approach might involve changing one hyperparameter at a time, but this ignores the interactions between parameters, such as the relationship between learning rate and batch size. More robust approaches include grid search and random search. Grid search involves defining a set of possible values for each hyperparameter and evaluating every possible combination. While exhaustive, this method is computationally expensive and inefficient when the search space is high-dimensional. Random search, by contrast, samples hyperparameter combinations randomly from a predefined distribution. Empirical evidence suggests that random search is often more efficient than grid search, as it allows for a broader exploration of the search space and can discover good configurations with fewer trials.

\medskip\noindent\emph{[Intermediate paragraphs omitted for length.]}

Furthermore, the reproducibility of experiments is a critical component of methodology. Because optimization algorithms like stochastic gradient descent involve random initialization and stochastic sampling, different runs with the same hyperparameters can yield different results. Therefore, it is standard practice to report the mean and variance of performance metrics over multiple random seeds. This ensures that the reported performance is not an artifact of a fortunate initialization. In this regard, the use of fixed random seeds for initialization can make debugging easier, while the use of multiple seeds provides a more robust assessment of the model's expected performance.
\endgroup